\documentclass[graybox]{svmult}

\usepackage{mathptmx}       
\usepackage{helvet}         
\usepackage{courier}        
\usepackage{type1cm}        
\usepackage{makeidx}         
\usepackage{graphicx}        
\usepackage{multicol}        
\usepackage[bottom]{footmisc}

\usepackage{amsmath}
\usepackage{amsfonts}
\usepackage{subfigure}

\usepackage{floatflt}
\usepackage{float}

\newcommand{\skipping}[1]{}

\usepackage{standalone}
\usepackage{tikz}
\usepackage{pgfplots}

\makeindex             

\begin{document}

\title*{Direct Variational Perspective Shape from Shading 
	    with Cartesian Depth Parametrisation}
\titlerunning{Direct Variational Perspective SfS 
	          with Cartesian Depth Parametrisation}
\author{Yong Chul Ju, 
	    Daniel Maurer, 
	    Michael Breu{\ss} and 
	    Andr{\'e}s Bruhn}
\institute{Yong Chul Ju \at 
	       Institute for Visualization and Interactive Systems\\
           University of Stuttgart, Germany\\
           \email{ju@vis.uni-stuttgart.de}
      \and Daniel Maurer \at 
           Institute for Visualization and Interactive Systems\\
           University of Stuttgart, Germany\\
           \email{maurer@vis.uni-stuttgart.de}
      \and Michael Breu{\ss} \at 
           Applied Mathematics and Computer Vision Group\\
           Brandenburg University of Technology Cottbus-Senftenberg, Germany\\
           \email{breuss@b-tu.de}
      \and Andr{\'e}s Bruhn \at 
           Institute for Visualization and Interactive Systems\\
           University of Stuttgart, Germany\\
           \email{bruhn@vis.uni-stuttgart.de}}

\maketitle

\abstract{Most of today's state-of-the-art methods for perspective shape from 
shading are modelled in terms of partial differential equations (PDEs) of 
Hamilton-Jacobi type. To improve the robustness of such methods w.r.t. noise 
and missing data, first approaches have recently been proposed that seek to 
embed the underlying PDE into a variational framework with data and 
smoothness term. So far, however, such methods either make use of a
radial depth parametrisation that makes the regularisation hard to interpret 
from a geometrical viewpoint or they consider indirect smoothness terms that 
require additional consistency constraints to provide valid solutions. 
Moreover the minimisation of such frameworks is an intricate task, since the
underlying energy is typically non-convex.
In this chapter we address all three of the aforementioned issues. 
First, we propose a novel variational model that operates directly 
on the Cartesian depth. In contrast to existing variational methods
for perspective shape from shading this refers to both the data and the 
smoothness term.
Moreover, we employ a direct second-order regulariser with edge-preservation 
property. This direct regulariser yields by construction valid 
solutions without requiring additional consistency constraints. 
Finally, we also propose a novel coarse-to-fine minimisation
framework based on an alternating explicit scheme. 
This framework allows us to avoid local minima during the minimisation 
and thus to improve the accuracy of the reconstruction.
Experiments show the good quality of our model as well as the usefulness
of the proposed numerical scheme.
}

\section{Introduction}
\label{sec:introduction}

Shape from Shading (SfS)\index{Shape from Shading} is a classic task in 
computer vision. Given information on light reflectance and illumination in a 
photographed scene, the aim of SfS is to compute based on the brightness
variation the 3D structure of a depicted object from a single input image. 
SfS has a wide variety of applications, ranging from large scale problems 
such as astronomy~\cite{Rindfleisch1966} or terrain 
reconstruction~\cite{Bors_TPAMI2003} to small scale tasks such as 
dentistry~\cite{Abdelrahim_BMVC2011} 
or endoscopy~\cite{Okatani_CVIU1997,Wang_MST2009,Wu_IJCV2010}.

\medskip

\noindent {\bf Classical Methods.}
First approaches to SfS go back to 1951 and 1966, respectively, when Van 
Diggelen \cite{VanDiggelen1951} and Rindfleisch~\cite{Rindfleisch1966} used 
SfS techniques to reconstruct the surface of the moon. 
Later on in the 1970's, Horn~\cite{Ho70} was the first one to tackle 
the SfS problem by solving a partial differential equation (PDE) approach. 
In 1981, he and Ikeuchi were also the first ones to model the SfS 
problem using a variational framework \cite{Ikeuchi_AI1981}. The most 
prominent classical variational approach is given by the work of Horn and 
Brooks \cite{HB86}. Assuming a simple {\em orthographic} projection model, 
a light source at {\em infinity} as well as a {\em Lambertian} reflectance 
model, they proposed to compute the normals of the unknown surface as 
minimiser of an energy functional.

Those first approaches, however, had several drawbacks. 
The model assumptions were {\em very simple} and mainly suitable 
in the context of astronomical applications. In fact, the use of an 
orthographic projection model with a light source located at infinity
requires the distances between camera, light source and illuminated 
object to be huge.
Also the {\em depth was not estimated directly} such that the SfS
process required a postprocessing step that performed a numerical 
integration of the estimated surface normals. Thereby, 
{\em inconsistent gradient fields} turned out to be a problem, so that 
extensions of the original model were required that tried to enforce this 
consistency during or after the estimation \cite{Frankot_TPAMI1988,HB89}. 
Finally, in case of variational methods, the smoothness term was restricted 
to a quadratic regulariser \cite{HB86,Ikeuchi_AI1981}. 
While such standard smoothness terms simplify the
minimisation of the underlying energy, they
do not allow to preserve discontinuities in the depth and thus lead
to {\em oversmoothed solutions}~\cite{Ol91}. For a detailed review of 
most of the classical methods 
the reader is referred to \cite{DFS08,Ho86,HB89,ZTCS99}.

\medskip

\noindent {\bf Perspective Shape from Shading.} At the end of the 1990's 
research mainly focused on novel concepts for formulating orthographic SfS
such as viscosity solutions \cite{RouyTourin_SINUM1992} 
and level set formulations \cite{Kimmel_IJCV1995}.
However, for most applications results were not satisfactory \cite{ZTCS99}. 
In the early 2000's, the situation changed completely. 
Inspired by the work of Okatani and Deguchi \cite{Okatani_CVIU1997},
independently, Prados and his co-workers \cite{PF03b,Prados_CVPR2005} 
as well as two other research groups \cite{CCDG04a,TSY03} proposed to 
consider a {\em perspective} camera model. Evidently, such a model is 
particularly appropriate for tasks that require the object to be relatively 
close to the camera such as e.g.\ in medical endoscopy. In such
cases the perspective effects dominate and an orthographic projection 
model would cause significant systematic errors as shown in \cite{TSY05}. 
Secondly, Prados and colleagues proposed to 
{\em shift the light source location from infinity to the camera centre} 
which can be seen as a good approximation of a camera with photoflash. 
This made shape from shading attractive for a variety of photo-based 
applications. Finally, also a physically motivated 
{\em light attenuation term} was introduced 
that models a quadratic fall-off due to the inverse square law. 
As discussed in \cite{BCDFV12}, the use of 
this term largely resolved the convex-concave ambiguity 
that was inherent to the classical orthographic model although some 
ambiguities are still present. Even the generalisation of such 
approaches to advanced reflectance models such as the Oren-Nayar 
\cite{Oren_IJCV1995} or the Phong reflectance model \cite{Phong_CACM1975} 
have been recently investigated \cite{AF2006,VBW08}.

However, this evolution of SfS models was accompanied by a different 
way of formulating the SfS problem. Instead of using variational methods, 
the perspective SfS problem was formulated in terms of
hyperbolic PDEs \cite{Prados_CVPR2005}. 
Although such PDE formulations allow for an efficient computation
of the solution using fast marching schemes \cite{Sethian1999}, they 
suffer from two inherent drawbacks: (i) On the one hand, they
are prone to {\em noise} and {\em missing data}, since they do not rely on any
form of regularisation or filling-in. This can be particularly 
problematic in the context of real-world images. (ii) On the other 
hand, it is {\em difficult to extend} the underlying model of such 
PDE-based schemes by additional constraints 
such as smoothness terms, multiple views, or additional light sources. 
While there have been recently some PDE-based approaches to
photometric stereo \cite{MF13},
one has to take care of ensuring the uniqueness of the solution if the
input data from multiple images is not consistent, cf.\ the discussion 
in \cite{Mecca_SIIMS14}.
  
\medskip 

\noindent 
{\bf Variational Perspective Shape from Shading.} Given the 
flexibility and robustness of variational methods, it is not
surprising that recently researchers tried to close the 
evolutionary loop by integrating the perspective SfS model 
into a suitable variational framework. So far, however, there
are only a few works in the literature that deal with this
recent idea. On the one hand, there 
is the work of Ju {\em et al.} 
\cite{JBB2014} that embeds the PDE of Prados {\em et al.} 
\cite{Prados_CVPR2005}
as data term into a variational model and complements it with
a discontinuity-preserving second order smoothness term. However, since 
the approach penalises deviations from the PDE directly and uses a 
parametrisation in terms of the radial depth, deviations in both the 
data and the smoothness term are 
difficult to interpret geometrically or photometrically. On the other hand, 
there is the approach of Abdelrahim {\em et al.} \cite{Abdelrahim_ICIP 2013} 
that formulates the data term in terms of brightness differences and makes 
use of a Cartesian depth parametrisation. 
While the corresponding energy functional
is thus more meaningful from a geometric and photometric viewpoint, 
it defines smoothness based on surface normals and thus needs an 
additional integrability constraint. 
Moreover, the corresponding smoothness term is restricted to a simple 
homogeneous regulariser that does not allow to preserve object edges 
during the reconstruction. 
Finally, there are the works of Zhang {\em et al.}
\cite{ZYT07} and Wu {\em et al.} \cite{Wu_IJCV2010} that also
make use of a Cartesian depth parametrisation but rely on
an indirect estimation using auxiliary variables. While
the approach of Zhang {\em et al.} \cite{ZYT07} resolves the
resulting consistency problem by considering an integrability constraint, 
the method of Wu {\em et al.} 
\cite{Wu_IJCV2010} repeatedly integrates the surface 
normals during computation to ensure valid solutions. Moreover, 
both approaches use derivations for their surface normals
that are based on the orthographic projection model of 
Horn and Brooks \cite{HB89}. 
Unfortunately, the resulting models are thus only valid
in case of weak perspective distortions.

A final issue that is common to all four of the aforementioned works 
is the difficulty of minimising the underlying energy. Since
this energy is non-convex, two of the methods rely on 
initialisations provided by closely related PDE-based SfS approaches 
\cite{Abdelrahim_ICIP 2013, ZYT07}. This, however, 
contradicts the idea of introducing robustness 
into the estimation -- in particular in the presence of noise
or missing data.  In contrast, the other two methods 
estimate the solution from scratch \cite{JBB2014,Wu_IJCV2010}.
However, those methods do not provide any quantitative assessment of
the reconstruction quality.
 
Let us summarise: While from a modelling viewpoint, it would be 
desirable to design a variational model that directly solves for 
the Cartesian depth without the need of integrability constraints 
or repeated integrations steps, it would be helpful from an 
optimisation viewpoint to develop a minimisation scheme that neither 
depends on the solution of other SfS techniques as in 
\cite{Abdelrahim_ICIP 2013,ZYT07} nor requires an accurate initialisation 
to produce meaningful results.

\medskip

\noindent 
{\bf Our Contributions.} In this book chapter we contribute to the field of
variational SfS in three ways:
(i) First, we consider a variational model for perspective 
SfS that makes use of a Cartesian depth parametrisation and an 
edge-preserving Cartesian depth regularisation. 
By penalising deviations from the image brightness 
in the data term and regularising the Cartesian depth in the smoothness 
term directly, we obtain an approach that is geometrically and 
photometrically meaningful. 
In this context, we also point out a popular mistake in the derivation 
of the surface normal and show two different ways to derive the 
normal correctly.
(ii) Our method is a direct approach to depth computation, i.e.\ 
it does not yield gradient fields that need to be integrated in a 
subsequent step, nor do we employ integrability constraints. 
(iii) Apart from the novel model, 
we also propose a novel minimisation strategy. By embedding an
alternating explicit scheme into a coarse-to-fine scheme, we obtain 
an optimisation framework that allows to obtain significantly better 
results than a traditional explicit scheme.
Experiments with synthetic and real-world images show the good quality of
our reconstructions and the advantages of our numerical scheme.

\medskip

\noindent {\bf Organisation of the Chapter.} 
In Section~2 we propose a novel PDE-based model for 
perspective SfS that is based on a Cartesian parametrisation 
of the depth. In Section~3 we then embed this PDE into a
variational framework with appropriate second order smoothness 
term. Details on the minimisation and the discretisation are
provided in Section~4, while Section~\ref{sec:intrinsic} 
comments on the integration of intrinsic camera parameters.
Finally, a detailed evaluation of our approach is presented in 
Section~\ref{sec:experimental-results}. The paper concludes with a summary 
in Section~\ref{sec:conclusion}.

\section{Perspective SfS with Cartesian Depth Parametrisation}
\label{sec:persp-shape-from}

In this section, we introduce a novel PDE-based SfS model that is 
parametrised in terms of the {\em Cartesian} depth. In contrast to 
most existing SfS models that estimate the radial depth or multiples
thereof, such a Cartesian parametrisation expresses 
the unknown surface directly in terms of the Euclidean distance 
along the $z$-axis, which is the axis orthogonal to the image plane. 

\begin{figure}[t!]
	\centering
	\includegraphics[width=0.6\linewidth]{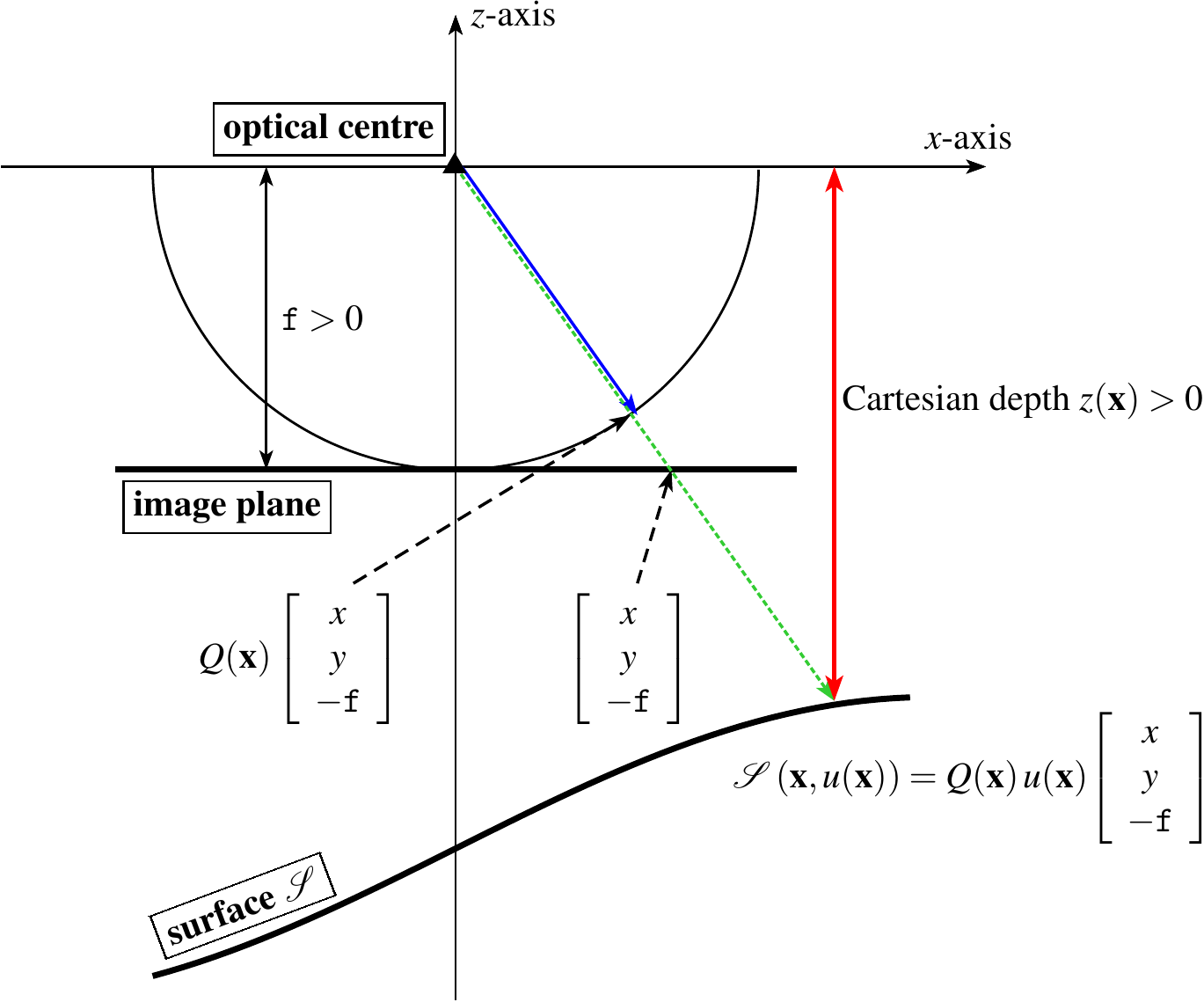} 
	\caption{ Relation between the radial depth factor $u(\mathbf{x})$ 
		(quotient between green and blue distance) that denotes the depth in 
		multiples of the focal length $\mathtt{f}$ and the Cartesian depth 
		$z(\mathbf{x})$ (red distance).} 
	\label{fig:parameterisation}
\end{figure}

\medskip

\noindent{\bf Parametrisation of the Surface.}
The starting point for our new model is formed by the classical 
PDE-approach of 
Prados {\em et al.} \cite{Prados_CVPR2005} which is originally parametrised 
in terms of the {\em radial} depth. Key assumptions of this SfS model are that 
a point light source is located at the {\em optical centre} of a perspective 
camera and that the surface reflectance is {\em Lambertian} with uniform 
albedo that is fixed to one.
The unknown surface 
$\mathcal{S}: \overline{\Omega}_{\mathbf{x}} \rightarrow \mathbb{R}^3$ 
can then be described as

\begin{equation}
  \label{eq:13}
  \mathcal{S} \left( \mathbf{x}, u(\mathbf{x}) \right)
  =
  \left\{ \dfrac{\mathtt{f}}{\sqrt{\left| \mathbf{x} \right|^2 + 
  	\mathtt{f}^2}} \; u(\mathbf{x})
    \left[ \left.
        \begin{array}{c}
          x \\
          y \\
          - \mathtt{f}
        \end{array}
      \right]
    \right| \; \mathbf{x}
    :=
    \left( x, y \right)^\top \in \overline{\Omega}_{\mathbf{x}} \right\} \, ,
\end{equation}
where $\mathbf{x}=(x, y)^\top \in \overline{\Omega}_{\mathbf{x}}$ is the 
position in the closure $\overline{\Omega}_{\mathbf{x}}$ of the rectangular 
image domain $\Omega_{\mathbf{x}} \subset \mathbb{R}^2$, $\mathtt{f}$ 
denotes the focal length of the camera and $u(\mathbf{x})$ is a multiple of 
$\mathtt{f}$ that describes the radial distance (depth) of the surface from 
the camera centre.

Since the third component in Eq. \eqref{eq:13} corresponds to the negative
Cartesian depth $z$,
we can derive the following 
relationship to the radial depth $u \, \mathtt{f}$
\begin{equation}
  \label{eq:64}
    z(\mathbf{x}) = 
    \dfrac{\mathtt{f}}{\sqrt{\left| \mathbf{x} \right|^2 + \mathtt{f}^2}} 
    \, u(\mathbf{x}) \, \mathtt{f}
    \; \stackrel{\mbox{\tiny \eqref{eq:13}}}{=} \; Q(\mathbf{x}) 
    \, u(\mathbf{x}) \, \mathtt{f} \, ,
\end{equation}
where $Q(\mathbf{x})$ denotes a spatially variant conversion factor given by
\begin{equation}
  \label{eq:12}
  Q(\mathbf{x}) = \dfrac{\mathtt{f}}{\sqrt{|\mathbf{x}|^2 + \mathtt{f}^2}} \, .
\end{equation}
This relation is illustrated in Figure \ref{fig:parameterisation}.

Plugging Eq. \eqref{eq:12} into Eq. \eqref{eq:13}, we then obtain the 
parametrisation of the original surface $\mathcal{S}$ with respect to the 
Cartesian depth $z$
\begin{equation}
  \label{eq:77}
  \mathcal{S} \left( \mathbf{x}, z(\mathbf{x}) \right)
  \stackrel{\mbox{\tiny \eqref{eq:13}}}{=} Q(\mathbf{x}) \, u(\mathbf{x}) 
  \left[
    \begin{array}{c}
      x \vspace{0.5ex}\\ 
      y \vspace{0.5ex}\\ 
      -\mathtt{f}
    \end{array}
  \right] 
  \stackrel{\mbox{\tiny \eqref{eq:64}}}{=}
  Q(\mathbf{x}) \, \left( \dfrac{z(\mathbf{x})}{\mathtt{f} 
  \, Q(\mathbf{x})} \right)
  \left[
    \begin{array}{c}
      x \vspace{0.5ex}\\ 
      y \vspace{0.5ex}\\ 
      -\mathtt{f}
    \end{array}
  \right] 
  =
  \left[
    \begin{array}{c}
      \dfrac{z(\mathbf{x}) \, x}{\mathtt{f}} \vspace{1ex}\\ 
      \dfrac{z(\mathbf{x}) \, y}{\mathtt{f}} \vspace{1ex}\\ 
      - z(\mathbf{x}) 
    \end{array}
  \right] 
\, .
\end{equation}

\noindent{\bf Brightness Equation.}
After we have parametrised the original surface in terms of the Cartesian 
depth, let us now derive the resulting brightness equation that relates the 
local orientation of the surface to the image brightness. 
Assuming a Lambertian reflectance model and a quadratic light
attenuation term that follows the inverse square law, we obtain
the following general brightness equation \cite{Prados_CVPR2005}:
\begin{equation}
  \label{eq:15}
  I(\mathbf{x}) 
  = 
  \dfrac{1}{r(\mathbf{x})^2} 
  \left( 
       \dfrac{\mathbf{n}( \mathbf{x})}{|\mathbf{n}( \mathbf{x})|} 
       \cdot \mathbf{L}(\mathbf{x})    
  \right) \, ,
\end{equation}
where $I$ is the recorded image, ${\mathbf{n}}
$ is the surface 
normal vector, $\mathbf{L}$ stands for the normalised light direction vector, 
and $r$ is the (radial) distance of the light source to the surface. 
Knowing that $r=\mathtt{f} \, u$
and using Eq. \eqref{eq:64} we can express the quadratic light attenuation term
using the Cartesian depth $z$
\begin{equation}
  \label{eq:99}
 r(\mathbf{x}) 
 = \mathtt{f} u(\mathbf{x}) 
 = \dfrac{z(\mathbf{x})}{Q(\mathbf{x})} \; \Rightarrow \;\;\;\; 
   \dfrac{1}{r(\mathbf{x})^2} 
 = \frac{Q(\mathbf{x})^2}{z(\mathbf{x})^2}\, .
\end{equation}
What remains to be computed in terms of the Cartesian depth are the 
surface normal $\mathbf{n}$ and the 
light direction vector $\mathbf{L}$, respectively.

\medskip
\noindent {\bf Surface Normal.}
Let us start by deriving the surface normal.
Since the surface normal is the normal vector of the tangent plane, we first
have to compute the partial derivatives of the surface in Eq. \eqref{eq:77} 
in $x$- and $y$-direction, respectively

\begin{equation}
\mathcal{S}_{x}( \mathbf{x}, z )  = 
\left[
  \begin{array}{c}
  \dfrac{z_{x} \, x + z}{\mathtt{f}} \vspace{1ex}\\ 
  \dfrac{z_{x} \, y}{\mathtt{f}} \vspace{1ex}\\ 
  - z_{x}
  \end{array}
\right]
\; ,
\;\;\;\;\;
\mathcal{S}_{y}( \mathbf{x}, z ) = 
\left[
  \begin{array}{c}
  \dfrac{z_{y} \, x}{\mathtt{f}} \vspace{1ex}\\ 
  \dfrac{z_{y} \, y + z}{\mathtt{f}} \vspace{1ex}\\ 
  - z_{y}
  \end{array}
\right] \, .
\end{equation}
\pagebreak

\noindent
Here and for the whole paper we dropped the spatial dependency of $z$, $z_x$ 
and $z_y$ on $\mathbf{x}$ for the sake of clarity. Taking the cross-product 
then yields the direction of the surface normal
\begin{equation}
  \label{eq:17}
  \mathbf{n}( \mathbf{x} )
  =
  \mathcal{S}_{x}( \mathbf{x}, z ) \times \mathcal{S}_{y}( \mathbf{x}, z )  
  =
  \left[
  \begin{array}{c}
    \dfrac{z_{x} \, z}{\mathtt{f}} \vspace{1ex}\\
    \dfrac{z_{y} \, z}{\mathtt{f}} \vspace{1ex}\\
    \dfrac{z \, \left[ \left( \nabla z \cdot \mathbf{x} \right) 
    + z \right]}{\mathtt{f}^2}
  \end{array}
  \right] \, .
\end{equation}

\medskip

\noindent {\bf Light Direction.}
Let us now turn towards the computation of the light direction. 
Since the light source is assumed to be located in the camera centre 
which coincides with the origin of the coordinate system,
the direction of the light rays and the direction of the optical 
rays coincide (up to sign). 
Hence, the light direction can just be read off Eq. \eqref{eq:13} as
\begin{equation}
  \label{eq:18}
  \mathbf{L}(\mathbf{x})
  =
  \dfrac{1}{\sqrt{|\mathbf{x}|^2 + \mathtt{f}^2}}
  \left[
    \begin{array}{c}
    -{x} \\      
    -{y} \\      
    \mathtt{f}
    \end{array}
  \right] \, .
\end{equation}

\medskip

\noindent {\bf PDE-Based Model}.
By plugging the surface normal from Eq. \eqref{eq:17} and the light direction 
from Eq. \eqref{eq:18} into the brightness equation \eqref{eq:15}
we finally obtain our perspective SfS model with the new Cartesian depth
parametrisation
\begin{equation}
\label{eq:16}
I
-
\dfrac{Q^3}{z \, \sqrt{ \mathtt{f}^2 \, \left| \nabla z  \right|^2
		+ \left[ \left( \nabla z  \cdot \mathbf{x} \right)
		+
		z  \right]^2 } } 
= 0 \, .
\end{equation}

Here and for the whole paper we dropped the spatial dependency of $I$ and 
$Q$ on $\mathbf{x}$ for the sake of clarity.

The main properties of our new model \eqref{eq:16} 
are naturally inherited from the original PDE \cite{Prados_CVPR2005}:
(i) Eq. \eqref{eq:16} still belongs to the class of Hamilton-Jacobi 
equations (HJEs) which have been intensively studied in the SfS 
literature. (ii)
Therefore, well-posedness can be achieved in the viscosity sense 
\cite{BCDFV12, Crandall_TAMS1983, Prados_CVPR2005}.
(iii) Proper numerical discretisations must be considered 
when solving the HJE. 

Let us note that the framework of viscosity solutions is a natural setting 
for HJEs such as Eq.\ \eqref{eq:16}. The basic idea behind the notion of 
viscosity solutions is to add a (typically, second order) regularisation 
term to the PDE and study the solution as this term goes to zero. 
This proceeding yields desirable stability properties and enables
to consider even solutions with non-differentiable features like e.g.\ kinks.
We refer the interested reader to \cite{Barles94, Crandall_TAMS1983} 
for studying properties of viscosity solutions and to \cite{CP14} for their 
use in computer vision.

Furthermore, please note that our model can be seen as a generalisation of the 
PDE-based approach in \cite{ZYT07b} that already makes use of the Cartesian 
depth parametrisation, but does not yet consider the light attenuation term 
from physics.

\section{Variational Model for Perspective SfS 
	     with Cartesian Depth Parametrisation}
\label{sec:var-model}

So far we have derived a novel PDE-based model for perspective SfS 
with Cartesian depth parametrisation. Let us now discuss how this
model can be integrated into a variational framework with smoothness term. 

\medskip

\noindent {\bf Variational Model.}
To this end, we follow the idea from \cite{JBB2014} and 
use a quadratic error term based on our novel PDE as data term 
which is complemented with a suitable 
second order regulariser. More precisely, we propose to compute
the Cartesian depth $z$ as minimiser of the following energy
functional
\begin{equation}
  \label{eq:2}
  E \left( z \right)
   = \int_{\overline{\Omega}_{\mathbf{x}}} \, c(\mathbf{x}) \;
   \underbrace{ 
     D(\mathbf{x} , z, \nabla z) }_{\mbox{Data term}}
         + \,
         \; \alpha \;  
         \underbrace{ S(\mbox{Hess}(z)) }_{\!\!\!\!\!\!\!\!\!\!\!\!\!\mbox{Smoothness term}\!\!\!\!\!\!\!\!\!\!\!\!\!}
         \, \mathrm{d}\mathbf{x} \; ,
\end{equation}
where $D$ is the data term, $S$ is the smoothness term, 
$c: \mathbf{x} \in \overline{\Omega}_{\mathbf{x}} \subset \mathbb{R}^2 \rightarrow [0,1]$ 
is a confidence function and $\alpha \in \mathbb{R}^+$ 
is a regularisation parameter that steers the degree of smoothness of the 
solution. As mentioned before our data term is based on a quadratic 
formulation that penalises deviations from our novel PDE. It is given by
\begin{equation}
  \label{eq:3}
  D( \mathbf{x}, z, \nabla z )
  =
  \left(I(\mathbf{x}) - \dfrac{Q(\mathbf{x})^3}{z \,\, W( \mathbf{x}, z, \nabla z ) } \right)^2
\end{equation}
with
\begin{equation}
  \label{eq:4}
  W( \mathbf{x}, z, \nabla z )
  =
  \sqrt{ \mathtt{f}^2 \, \left| \nabla z  \right|^2
    + \left[ \left( \nabla z  \cdot \mathbf{x} \right)
      +
      z \right]^2 } \; .
\end{equation}
As smoothness term, we propose to use the following subquadratic 
and thus edge-preserving second-order 
regulariser based on the Frobenius norm of the Hessian
\begin{equation}
  \label{eq:5}
   S \left( \mbox{Hess}(z) \right) =    
  \Psi \left( \left\| \mbox{Hess}(z) \right\|^2_{\tiny F} \right)
  =
  \Psi \left(
  z_{xx}^2
  +
  2 z_{xy}^2 
  +
  z_{yy}^2
  \right) 
\end{equation}
where $\Psi$ is the Charbonnier 
function~\cite{Charbonnier_TIP1997}
\begin{equation}
  \label{eq:11}
  \Psi(s^2) = 2 \lambda^2 \sqrt{ 1 + \tfrac{s^2}{\lambda^2} }
\end{equation}
with contrast parameter $\lambda$. Such higher-order smoothness terms have 
already been
successfully applied in the context of perspective SfS  parametrised in 
terms of the radial depth \cite{JBB2014}, 
orthographic SfS \cite{Vogel_SSVM2007}, 
image denoising \cite{Lysaker_TIP2003}, optical lithography 
\cite{Estellers_TIP2014} and motion estimation \cite{Demetz_ECCV2014}.
Finally, the use of the confidence function $c$ in the data term allows 
to exclude unreliable image regions  which have been identified a priori, 
e.g.\ 
by a texture detector or by a background segmentation algorithm. 
Such functions are particularly useful in the context of real-world images 
that contain texture, noise, or missing data \cite{CCDG08,JBB2014}.

\begin{table}[t!]
\caption{Comparison of the literature on variational models for 
	     perspective shape from shading.}
\label{tab:methods}
\begin{center}
\begin{tabular}{|clc||ccc|ccc|ccc|ccc|ccc|} \hline
& &&& &&&  &&&  &&&  &&& &\\[-2mm]
& &&& Zhang {\em et al.} &&&  Wu {\em et al.} &&& Abdelrahim {\em et al.} &&& Ju {\em et al.} &&& {\bf Our} &\\
& &&& \cite{ZYT07} &&&  \cite{Wu_IJCV2010} &&& \cite{Abdelrahim_ICIP 2013} &&& \cite{JBB2014} &&& {\bf Work} &\\[-2mm]
& &&& &&&  &&&  &&&  &&& &\\\hline\hline
& &&& &&&  &&&  &&&  &&& &\\[-2mm]
& &&& Cartesian &&&  Cartesian &&& Cartesian  &&& radial &&& Cartesian &\\[-2mm]
& Parametrisation &&& &&& &&& &&& &&& &\\[-2mm]
& &&& depth &&& depth &&& depth &&& depth &&& depth &\\[-2mm]
& &&& &&&  &&&  &&&  &&& &\\\hline
& &&& &&&  &&&  &&&  &&& &\\[-2mm]
& Reprojection Error &&& &&&  &&&  &&&  &&& &\\[-2mm]
&  &&& $\checkmark$ &&& $\checkmark$ &&& $\checkmark$ &&& $-$ &&& $\checkmark$ &\\[-2mm]
& as Data Term &&& &&& &&& &&& &&& &\\[-2mm]
& &&& &&&  &&&  &&&  &&& &\\\hline
& &&& &&&  &&&  &&&  &&& &\\[-2mm]
& Light Attenuation Factor &&& $-$ &&& $\checkmark$ &&& $\;\;\;\checkmark^{\; 1}$ &&& $\checkmark$ &&& $\checkmark$ &\\[-2mm]
& &&& &&&  &&&  &&&  &&& &\\\hline
& &&& &&&  &&&  &&&  &&& &\\[-2mm]
& Correct Surface Normal &&& $\;\;-^{\; 2}$ &&& $\;\;-^{\; 2}$ &&& $\;\;\;\checkmark^{\; 3}$ &&& $\checkmark$ &&& $\checkmark$ &\\[-2mm]
& &&& &&&  &&&  &&&  &&& &\\\hline
& &&& &&&  &&&  &&&  &&& &\\[-2mm]
& &&& Cartesian &&&  Cartesian &&& Cartesian  &&& radial &&& Cartesian &\\[-2mm]
& Regularisation &&& &&& &&& &&& &&& &\\[-2mm]
& &&& depth &&& depth &&& surface normal &&& depth &&& depth &\\[-2mm]
& &&& &&&  &&&  &&&  &&& &\\\hline
& &&& &&&  &&&  &&&  &&& &\\[-2mm]
& Edge Preservation  &&& $-$ &&& $-$ &&& $-$ &&& $\checkmark$ &&& $\checkmark$ &\\[-2mm]
& &&& &&&  &&&  &&&  &&& &\\\hline
& &&& &&&  &&&  &&&  &&& &\\[-2mm]
& No Integrability Term &&& $-$ &&& $\;\;-^{\; 4}$ &&& $-$ &&& $\checkmark$ &&& $\checkmark$ &\\[-2mm]
& &&& &&&  &&&  &&&  &&& &\\\hline
& &&& &&&  &&&  &&&  &&& &\\[-2mm]
& Direct Estimation $^5$ &&& $-$ &&& $-$ &&& $\checkmark$ &&& $\checkmark$ &&& $\checkmark$ &\\[-2mm]
& &&& &&&  &&&  &&&  &&& &\\\hline

\end{tabular}
\end{center}
\vspace{-1mm}
$^1$ factor not expressed in terms of the Cartesian depth\\
$^2$ see explanation in appendix\\
$^3$ no details given in the paper but derivations shown in \cite{Ab14}\\
$^4$ integrability constraint realised via repeated integration of surface normals\\
$^5$ depth is computed without extra variables for surface normals\\
\end{table}

\pagebreak
\noindent {\bf Properties.}
Our variational model from  Eq. \eqref{eq:2} has the following distinct 
features: 
\begin{enumerate}
	\item[(i)] Since the data term in Eq. \eqref{eq:3} is inherited from 
	Eq. \eqref{eq:16},
	the perspective camera projection is already taken into account.
	Moreover, since the reprojection error is penalised in the data term,
	deviations have a {\em photometric} interpretation.
	\item[(ii)] Since the regulariser is applied directly to the Cartesian
	depth, also deviations from smoothness become now more meaningful
	than in the case of a radial depth parametrisation. In particular,
	they can be interpreted {\em geometrically}. 
	\item[(iii)] Moreover, in contrast
	to most existing approaches, the regulariser is able to 
	{\em preserve edges} in the reconstruction despite of the regularisation effect.
	\item[(iv)] Unreliable regions can be excluded from the data term via a 
	{\em confidence function} such that the smoothness term takes over and 
	fills in information from the neighbourhood.
	This can be advantageous in the context of texture, noise, or missing data.
	Please note that in contrast to \cite{JBB2014}, we always guarantee a fixed 
	amount of regularisation by not restricting the smoothness term 
	to unreliable locations.
	\item[(v)] The depth of the surface is {\em directly} computed
	since we minimise for the unknown depth $z$ in Eq. \eqref{eq:2}. 
	This is in contrast to most variational methods that
	estimate the depth in two steps, see e.g. \cite{BH85, Frankot_TPAMI1988, Ikeuchi_AI1981} 
	where first the surface normals are computed by a variational model
	and then the depth is determined by integration.
	\item[(vi)] The solution given by the model fulfils the 
	{\em integrability constraint} per construction since we solve 
	for $z$ and use $z_{xy} = z_{yx}$ in
	the smoothness term. Otherwise, such as in
	\cite{Abdelrahim_ICIP 2013}, an additional integrability term
	would be needed to encourage valid solutions.
	\item[(vii)] Another advantage of the new parametrisation is that it
	allows a straightforward {\em combination} with other reconstruction 
	methods such as stereo \cite{Robert_ECCV1996} 
	or scene flow estimation \cite{Basha_IJCV2012}, 
	since such approaches typically make use of the same Cartesian 
	parametrisation and thus could be easily integrated into a joint framework. 
	\item[(viii)] A final advantage is the fact that the approach could easily be extended
	to {\em multiple views}, since transformations between the views
	are simpler if the approach is parametrised in terms of the Cartesian depth
	instead of the radial depth. 
\end{enumerate}
To make the difference of our model to other variational approaches 
from the literature explicit, the features of the different methods are compared in 
Table \ref{tab:methods}.

\vspace{-3mm}

\section{Minimisation}
\label{sec:minimisation}

\vspace{-1mm}

Let us now discuss the minimisation of the proposed energy. To this end,
we will first derive the associated Euler-Lagrange equation and then
discuss its discretisation. Finally, we will sketch a coarse-to-fine 
minimisation strategy with an alternating explicit scheme to solve the 
resulting nonlinear equations.

\medskip

\noindent {\bf Euler-Lagrange Equation}.
The calculus of variations \cite{CourantHilbert1953} tells us that the
minimiser $z$ of our energy in Eq. \eqref{eq:2} has to fulfil
the corresponding Euler-Lagrange equation. Omitting the dependencies on all variables in order to ease the readability, this equation is given by
\begin{eqnarray}  
      0 & \; = \; & \left[ cD + \alpha S \right]_z 
      - \dfrac{\partial}{\partial x} \left[ cD + \alpha S \right]_{z_{x}} 
      - \dfrac{\partial}{\partial y} \left[ cD + \alpha S \right]_{z_{y}} \\[-1mm]
      & & \; + \dfrac{\partial^2}{\partial x^2} \left[ cD + \alpha S \right]_{z_{xx}}
      + 2 \dfrac{\partial^2}{\partial x \partial y} \left[ cD + \alpha S \right]_{z_{xy}}  
      + \dfrac{\partial^2}{\partial y^2} \left[ cD + \alpha S \right]_{z_{yy}} \nonumber  \\[0mm]      
      & = \; &      
      \left[ cD \right]_z      
      - \dfrac{\partial}{\partial x} \left[ cD \right]_{z_{x}}
      - \dfrac{\partial}{\partial y} \left[ cD \right]_{z_{y}}       
      +  \dfrac{\partial^2}{\partial x^2} 
      \overbrace{
      	\left[ cD \right]_{z_{xx}} }^{\equiv \, 0}      
      + 2   \dfrac{\partial^2}{\partial x \partial y} 
      \overbrace{
      	\left[ cD \right]_{z_{xy}} }^{\equiv \, 0}     
      +  \dfrac{\partial^2}{\partial y^2} 
      \overbrace{ \left[ cD \right]_{z_{yy}} }^{\equiv \, 0}  
      \nonumber \\[-1mm]
      & & \;       
      +  \underbrace{ \vphantom{\left[ cD \right]_{z_{yy}}} \left[ \alpha S \right]_z}_{\equiv \, 0}
      -  \dfrac{\partial}{\partial x} 
         \underbrace{ \vphantom{\left[ cD \right]_{z_{yy}}} \left[ \alpha S \right]_{z_{x}} }_{\equiv \, 0}
      -  \dfrac{\partial}{\partial y} 
         \underbrace{\left[ \alpha S \right]_{z_{y}} }_{\equiv \, 0}
      + \dfrac{\partial^2}{\partial x^2} \left[ \alpha S \right]_{z_{xx}}         
      + 2 \dfrac{\partial^2}{\partial x \partial y} \left[ \alpha S \right]_{z_{xy}}       
      + \dfrac{\partial^2}{\partial y^2} \left[ \alpha S \right]_{z_{yy}}      
      \nonumber \\[2mm]
      & = \; &
       c \; \left( \left[ D \right]_z
      - \dfrac{\partial}{\partial x} \left[ D \right]_{z_{x}}
      - \dfrac{\partial}{\partial y} \left[ D \right]_{z_{y}} \right) 
      + \dfrac{\partial^2}{\partial x^2} \left[ \alpha S \right]_{z_{xx}} 
      + 2 \dfrac{\partial^2}{\partial x \partial y} \left[ \alpha S \right]_{z_{xy}} 
      + \dfrac{\partial^2}{\partial y^2} \left[ \alpha S \right]_{z_{yy}} 
      \nonumber      \, ,
      \label{eq:49}
\end{eqnarray}
\vspace{1mm}

\noindent where we exploited the fact that
\begin{equation}
  \label{eq:28}
  \dfrac{\partial^2}{\partial x \partial y} \left[ cD + \alpha S \right]_{z_{xy}}
  =
  \dfrac{\partial^2}{\partial y \partial x} \left[ cD + \alpha S \right]_{z_{xy}} \, .
\end{equation}
On a structural level, this Euler-Lagrange equation is somewhat more complicated
than its counterparts for indirect methods in \cite{Wu_IJCV2010,ZYT07}. 
Such indirect methods model the surface normal using auxiliary variables $p=z_{x}$ and 
$q=z_{y}$ and thus do not have the additional data term contributions 
$\tfrac{\partial}{\partial x} \left[ D \right]_{z_{x}}$ 
and $\tfrac{\partial}{\partial y} \left[ D \right]_{z_{y}}$.

Let us now take a closer look at all the individual terms that occur in Eq. (\ref{eq:49}).
After some computations we obtain
\begin{eqnarray}
  \label{eq:1} 
     \left[ D \right]_z
     &=& 
     2 \, \left(  I - \dfrac{Q^3}{z \, W} \right) \,
     \left(
     \dfrac{Q^3}{z^2 \, W}
     +
     \dfrac{Q^3 }{z \, W^2} \left[ W \right]_z
     \right) \\
     &=& \nonumber
     2 \, \left(  I - \dfrac{Q^3}{z \, W} \right)
     \,
     \dfrac{Q^3}{z \, W}
     \,
     \left(
     \dfrac{1}{z} 
     +
     \dfrac{ \nabla z \cdot \mathbf{x} + z }{W^2}
     \right) 
\end{eqnarray}
\begin{eqnarray}
\label{eq:23}
\dfrac{\partial}{\partial x} \left[ D \right]_{z_{x}}
&=&
\left[
2 \, \left(  I - \dfrac{Q^3}{z \, W} \right) \,
\dfrac{Q^3}{z \, W^3} \,
\left[ W \right]_{x} \right]_{x}  \\
&=& \nonumber
\left[
2 \, \left(  I - \dfrac{Q^3}{z \, W} \right) \,
\dfrac{Q^3}{z \, W^3} \,
\left(
\mathtt{f}^2 \, z_{x} 
+
\left[ \nabla z \cdot \mathbf{x} + z \right] \, x
\right) \right]_{x}  
\end{eqnarray}
\begin{eqnarray}
  \label{eq:27}   
  \dfrac{\partial}{\partial y} \left[ D \right]_{z_{y}}
  &=& 
  \left[
  2 \, \left(  I - \dfrac{Q^3}{z \, W} \right) \,
  \dfrac{Q^3}{z \, W^3} \,
  \left[ W \right]_{y} \right]_{y}  \\
  &=& \nonumber 
  \left[
  2 \, \left(  I - \dfrac{Q^3}{z \, W} \right)
  \dfrac{Q^3}{z \, W^3} \,
  \left(
  \mathtt{f}^2 \, z_{y} 
  +
  \left[ \nabla z \cdot \mathbf{x} + z \right] \, y
  \right) \right]_{y}  
\end{eqnarray}
as well as
\begin{equation}
  \label{eq:29}
  \dfrac{\partial^2}{\partial x^2} \left[ S \right]_{z_{xx}} = 2 \, \dfrac{\partial^2}{\partial x^2} \left[ \Psi'(\left\| \mbox{Hess}(z) \right\|_F^2) \; z_{xx} \right]
, \quad
\end{equation}
\begin{equation}
  \label{eq:30}
  2 \dfrac{\partial^2}{\partial xy} \left[ S \right]_{z_{xy}} = 4 \, \dfrac{\partial^2}{\partial xy} \left[ \Psi'(\left\| \mbox{Hess}(z) \right\|_F^2) \; z_{xy} \right]
, \quad
\end{equation}
\begin{equation}
  \label{eq:31}
  \dfrac{\partial^2}{\partial y^2} \left[ S \right]_{z_{yy}} = 2 \, \dfrac{\partial^2}{\partial y^2} \left[ \Psi'(\left\| \mbox{Hess}(z) \right\|_F^2) \; z_{yy} \right]
\, ,
\end{equation}

\vspace{3mm}
\noindent where the derivative of the penaliser function $\Psi(s^2)$ reads
\begin{equation}
  \label{eq:80}
  \Psi'(s^2) = \frac{\partial}{\partial (s^2)} \Psi(s^2)= \frac{1}{\sqrt{ 1 + \tfrac{s^2}{\lambda^2}}} \; .
\end{equation}
While the contributions of the data term are related to the influence of $z$ and $\nabla z$ on the brightness 
equation, the contributions of the smoothness term define an edge-preserving fourth-order diffusion process. 
This becomes explicit as follows: Since $\Psi'(s^2)$ becomes small for large values of $s^2$, 
this reduces the effect of the smoothing at locations with high curvature, i.e. where 
$\left\| \mbox{Hess}(z) \right\|_F^2$ is large. 
After we have derived the resulting Euler-Lagrange equation, let us now discuss how this equation can be discretised appropriately.

\medskip

\noindent {\bf Discretisation.}
In order to discretise the contributions of the data term given by Eqs. \eqref{eq:1} -- \eqref{eq:27},
we employ the upwind scheme from \cite{RouyTourin_SINUM1992} in 
view of the hyperbolic nature of the underlying PDE. In 1D, the corresponding upwind discretisation reads 

\begin{equation}
  \label{eq:7}
  \widetilde{z}_{x} \approx \max \left( D^- z, - D^+ z, 0 \right) \, ,  
\end{equation}
with
\begin{equation}
  \label{eq:8}
  D^- z = \dfrac{z_i - z_{i-1} }{h_{x}}  
  \qquad \mbox{and} \qquad
  D^+ z = \dfrac{z_{i+1} - z_i }{h_{x}}  \, ,
\end{equation}
where $h_{x}$ denotes the grid size. 
Please note that in contrast to upwind schemes for eikonal equations \cite{Sethian1999}
that typically approximate only the magnitude of the gradient, 
the sign matters in our case, such that we have to choose 
\begin{equation}
	\label{eq:z_x}
	z_x = \left\{ 
	\begin{array}{lcl}
	D^+ z & & \mbox{if} \quad \widetilde{z}_x = -D^+ z \, , \\
	\widetilde{z}_x & & \mbox{otherwise} \, .
	\end{array}
	\right.
\end{equation}
This selects the actual forward difference, if the second argument in 
(\ref{eq:7}) is the maximum \cite{BCDFV10,BCDFV12}. This scheme can be 
extended in a straightforward way to 2D.
For discretising the contributions of the smoothness term, a standard central difference 
scheme is used. 

Since it is difficult to discretise the Euler-Lagrange equation directly,
we followed a {\em first discretise then optimise} scheme.
To this end, we used the aforementioned finite difference approximations
to discretise the energy in \eqref{eq:2} applying the upwind scheme for 
the data term and a central difference approximation for the smoothness term.
Then, by computing the derivatives of the discrete energy
we obtain a proper discretisation for the Euler-Lagrange equation.

Finally, by using the Euler forward time discretisation method
\begin{equation}
  \label{eq:20}
  z_t
  \approx
  \dfrac{z^{n+1} - z^n}{\tau} \, ,
\end{equation}
with $\tau$ being a time step size,
we can reformulate the solution of Eq. \eqref{eq:49} as the steady state of the corresponding evolution equation 
in artificial time. Thus we obtain the following explicit scheme 
\begin{equation}
  \label{eq:21}
  \dfrac{z^{n+1} - z^n}{\tau} + EL^n = 0
  \qquad
  \Leftrightarrow
  \qquad
  z^{n+1} = z^n - \tau \, EL^n \, ,
\end{equation}
where $EL^n$ is the discretisation of the Euler-Lagrange equation evaluated at time $n$.
Please note that this discretisation may change over time, since we re-discretised the energy in each iteration by adapting the direction of the discretisation of the upwind scheme (forward, backward, no contribution) based on evaluating Eqs. \eqref{eq:7} -- \eqref{eq:z_x} for the result of the previous time step.
In that sense we use a {\em lagged} discretisation approach, where the discretisation is updated in each iteration. 

\medskip

\noindent {\bf Coarse-to-Fine Approach.} Since the underlying energy functional
is highly non-convex, the proposed explicit scheme may get trapped in 
local minima. To tackle this problem, we propose to embed the estimation into 
a coarse-to-fine framework. Starting from a very coarse resolution, we successively 
refine the input image while repeatedly reconstructing the surface. Thereby, 
solutions from coarser levels serve as initialisation for the finer scales. 
Similar hierarchical schemes have already been successfully applied to 
many other problems in computer vision; see e.g.\ \cite{BCDFV10,Broxz_ECCV2004}. 

Apart from improving the quality of the results by avoiding local minima,
coarse-to-fine schemes also render the estimation more robust w.r.t.\ the choice of 
the initialisation. In fact, if sufficiently many resolution levels were used, 
we could hardly observe any 
impact of the initialisation on the quality of the final results. Since a good 
initial guess can still be useful to speed up the computation, we
propose to initialise the depth by pointwise solving 
the data term in Eq. \eqref{eq:3} for $\nabla z = 0$
\begin{equation}
  \label{eq:333}
  D( \mathbf{x}, z, 0 ) = 0 \;\;\;\; \Rightarrow \;\;\;\; z = \sqrt{\dfrac{Q(\mathbf{x})^3}{I(\mathbf{x})}} \, .
\end{equation}
This can be seen as an efficient compromise between using the full model 
which is evidently not feasible
and only considering the inverse square law, 
i.e. $z = 1/\sqrt{I(\mathbf{x})}$, which completely neglects the effect of the 
surface orientation and thus actually provides a local upper bound for the correct depth.
In any case, in contrast to other variational SfS methods from the literature, 
our technique does not have to rely on initialisations from
non-variational SfS approaches \cite{Abdelrahim_ICIP 2013,ZYT07} or 
surface integration methods \cite{Wu_IJCV2010}
to provide meaningful results.

Let us now discuss the details of our coarse-to-fine approach. To this end, we 
introduce the parameter $\eta$ that specifies the downsampling factor 
between two consecutive resolution levels and that is typically chosen in the 
interval $(0.5,1)$. Then the grid size at level $k$ of
our coarse-to-fine approach
can be computed as
\begin{eqnarray}
h^k_{x} = h_{x} \cdot \eta^{-k} \; , \;\;\;\;\; h^k_{y} = h_{y} \cdot \eta^{-k} \; .
\end{eqnarray}
where $k=0$ is the original resolution and $k=k_{\mathrm{max}}$ is the coarsest level.
This tells us that the grid size becomes larger at coarser scales which 
intuitively makes sense, since the size of the image plane remains constant
while the number of pixels decreases.
At the same time, however, this
increase of the grid size leads to a major problem: Since the contributions of the smoothness term
given by Eqs. \eqref{eq:29} -- \eqref{eq:31} involve fourth-order derivatives that scale proportionally 
to $1/h^4$, the strength of the regularisation actually decreases with $\eta^{4k}$ on coarser scales. 
In order to compensate for this effect, we thus propose to scale the smoothness weight $\alpha$ 
according to \begin{equation}
\alpha^k =  \eta^{-4k} \cdot \alpha \, .
\end{equation}
This guarantees a similar amount of regularisation for all resolution levels.

\medskip

\noindent {\bf Alternating Explicit Scheme.}
Finally, we observed in our experiments that the terms in Eqs. \eqref{eq:23} and \eqref{eq:27} 
that refer to the influence of the depth gradient $\nabla z$ on the brightness equation 
require to select the time step size $\tau$ rather small. In particular, these terms do
not have a weighting parameter such as the smoothness term that can be adjusted
appropriately.
As a consequence, the minimisation typically needs several
thousands or even millions of iterations. To counter this problem, we propose the following 
alternating estimation strategy at each resolution level: For a fixed number of iterations $n$, instead of performing $n$ 
iterations using the original explicit scheme, we propose an alternating iterative scheme
that first does $n/2$ iterations with a simplified explicit scheme neglecting the two terms in Eqs. 
\eqref{eq:23} and \eqref{eq:27}, followed by $n/2$ iterations with the entire explicit scheme. 
Since the neglected terms are based on second-order derivatives
and the remaining terms did not strongly affect the convergence, 
we empirically found out that we can choose the 
time step size approximately $\min(h_{x}^{-2}, h_{y}^{-2})$ times larger for the first $n/2$ iterations
(given that $h_{x},h_{y}\ll 1$). In our experiments this leads to speed-ups of about one to four orders of magnitude. Moreover, in most cases, even the simplified scheme was 
sufficient to achieve excellent results. Thereby one should  note that, from a numerical 
viewpoint, the simplified scheme can be understood as an optimisation method for a series of 
energy functionals of type of Eq. \eqref{eq:2}, where the gradient $\nabla z$ is lagging and 
thus has no direct influence on the optimisation.

\vspace{-4.5mm}

\section{Intrinsic Parameters}
\label{sec:intrinsic}

\vspace{-1.5mm}

So far we have derived a variational model for perspective SfS 
with Cartesian depth parametrisation that is given in terms of 
{\em image coordinates}.
Let us now discuss how the model and the minimisation has to be adapted 
if we additionally consider the intrinsic camera parameters,
i.e. if we express the model in terms of {\em pixel coordinates}.\linebreak

\medskip

\noindent{\bf Coordinate Transformation.}	 
Let the corresponding calibration matrix be given by 
\begin{equation}
\label{eq:200}
K
=
\left[
 \begin{array}{ccc}
 \mathtt{f} / h_{x} & 0 & c_1\\
 0 & \mathtt{f} / h_{y} & c_2 \\
 0 & 0 & 1\\
 \end{array}
\right] \, .
\end{equation}
where $(c_1,c_2)^\top$ denotes the location of the focal point, 
and $h_{x}$ and $h_{y}$ 
is the grid size in $x$- and $y$-direction, respectively \cite{HZ04}. 
Knowing this matrix allows us to reformulate the image coordinates 
$\mathbf{x}=(x, y)^\top$ of our original model in terms of pixel 
coordinates $\mathbf{a}=(a,b)^\top$. The corresponding transformation 
is given by
\begin{equation}
\label{eq:201}
\left[
 \begin{array}{c}
 a\\
 b\\
 - 1\\
 \end{array}
\right]
=
K \;
\dfrac{1}{\mathtt{f}}
\left[
 \begin{array}{c}
 x\\
 y\\
 - \mathtt{f}\\
 \end{array}
\right]
\; \Rightarrow \; 
\left[
 \begin{array}{c}
 x\\
 y\\
 - \mathtt{f}\\
 \end{array}
\right]
=
\mathtt{f} \; K^{-1} 
\left[
 \begin{array}{c}
 a\\
 b\\
 - 1\\
 \end{array}
\right] \; ,
\end{equation}
where one has to take care that the image plane is at distance $\mathtt{f}$
of the camera centre. Plugging Eq. \eqref{eq:200} into  Eq. \eqref{eq:201}
then yields
\begin{eqnarray}
\label{eq:202}
\mathbf{x}(\mathbf{a}) 
=
\left[
\begin{array}{c}
x (a) \\
y (b)
\end{array}
\right]
=
\left[
\begin{array}{cc}
h_{x} & 0 \\
0 & h_{y}
\end{array}
\right]
\left[
\begin{array}{c}
a  \\
b 
\end{array}
\right]
-
\left[
\begin{array}{c}
h_{x} \, c_1 \\
h_{y} \, c_2
\end{array}
\right] \; .
\end{eqnarray}

\medskip

\noindent {\bf Variational Model.}
Now we are in the position to reformulate our entire model in terms of pixel coordinates. Substituting Eq. \eqref{eq:202} into our 
original energy and transforming the integration domain $\overline{\Omega}_{\mathbf{a}}=\mathbf{x}^{-1}(\overline{\Omega}_{\mathbf{x}})$ accordingly, 
we obtain the following variational model expressed in terms of pixel coordinates
\begin{equation}
  \label{eq:2b}
  E \left( z(\mathbf{x(a)}) \right)
   = \int_{\overline{\Omega}_{\mathbf{a}}} \, c(\mathbf{x(a)}) \;
   \underbrace{ 
     D(\mathbf{x(a)} , z(\mathbf{x(a)}), \nabla z(\mathbf{x(a)})) }_{\mbox{Data term}}
         + \,
         \alpha 
         \underbrace{ S(\mbox{Hess}(z)(\mathbf{x(a)})) }_{\!\!\!\!\!\!\!\!\!\!\!\!\!\mbox{Smoothness term}\!\!\!\!\!\!\!\!\!\!\!\!\!}
         \, \mathrm{d}\mathbf{a} \; . \\[2mm]
\end{equation}

\noindent Please note that we omitted the substitution factor given by 
$|\mathrm{det}(J(\mathbf{x}(\mathbf{a})))|$, where $J$ 
is the Jacobian,
since this factor is constant 
and thus does not change the minimiser 
of our energy. Let us now derive the corresponding Euler-Lagrange equation for our novel model 
expressed in terms of pixel coordinates.

\medskip

\noindent {\bf Euler-Lagrange Equation.} Analogously to Eq. \eqref{eq:49} we drop the dependencies on all variables and obtain the following Euler-Lagrange equation
\begin{eqnarray}  
  \label{eq:49b}    
     0 & \; = \; &
       c \left( \left[ D \right]_z
      - \dfrac{\partial}{\partial a} \left[ D \right]_{z_{a}}
      - \dfrac{\partial}{\partial b} \left[ D \right]_{z_{b}} \right) \nonumber\\
      & & \; + \; \alpha \; \left( \dfrac{\partial^2}{\partial a^2} \left[ S \right]_{z_{aa}} 
      + 2 \dfrac{\partial^2}{\partial a \partial b} \left[ S \right]_{z_{ab}} 
      + \dfrac{\partial^2}{\partial b^2} \left[ S \right]_{z_{bb}} \right)           \\
      & \; = \; &
       \label{eq:49c}    
       c \left( \left[ D \right]_z
      - \dfrac{\partial}{\partial x} \left[ D \right]_{z_{x}}
      - \dfrac{\partial}{\partial y} \left[ D \right]_{z_{y}} \right) \nonumber\\
      & & \; + \; \alpha \; \left( \dfrac{\partial^2}{\partial x^2} \left[ S \right]_{z_{xx}} 
      + 2 \dfrac{\partial^2}{\partial x \partial y} \left[ S \right]_{z_{xy}} 
      + \dfrac{\partial^2}{\partial y^2} \left[ S \right]_{z_{yy}} \right)   \; ,        
\end{eqnarray}
where we exploited the following relation between derivatives in pixel and image coordinates 
due to Eq. \eqref{eq:202}
\begin{eqnarray}
\label{eq:203}
&&\dfrac{\partial}{\partial a}  =  h_{x} \; \dfrac{\partial}{\partial x} \; , \;\;\;\;\;
\dfrac{\partial}{\partial b}  =  h_{y} \; \dfrac{\partial}{\partial y} \; , \;\;\;\;\; 
\dfrac{\partial}{\partial z_{a}}  = \dfrac{1}{h_{x}} \; \dfrac{\partial}{\partial z_{x}}  \; , \;\;\;\;\;
\dfrac{\partial}{\partial z_{b}}  = \dfrac{1}{h_{y}} \; \dfrac{\partial}{\partial z_{y}}  \; , \;\;\;\;\; \\
&&
\dfrac{\partial}{\partial z_{aa}} = \dfrac{1}{h_{x} h_{x}} \; \dfrac{\partial}{\partial z_{xx}} \; , \;\;\;\;\;
\dfrac{\partial}{\partial z_{ab}} = \dfrac{1}{h_{x} h_{y}} \; \dfrac{\partial}{\partial z_{xy}} \; , \;\;\;\;\;
\dfrac{\partial}{\partial z_{bb}} = \dfrac{1}{h_{y} h_{y}} \; \dfrac{\partial}{\partial z_{yy}} 
\, .
\end{eqnarray}

The equality between Eq. \eqref{eq:49b} and Eq. \eqref{eq:49c} shows that 
the Euler-Lagrange equations of our models in pixel and image coordinates are 
basically identical. One only has to parametrise the terms $\eqref{eq:1}$--$\eqref{eq:31}$ 
that have been originally derived in image coordinates using the coordinate transform 
in Eq. \eqref{eq:202}.
Apart from that, the discretisation 
can be performed in accordance with our explanations from the 
previous section. In this context, the grid size is given by the intrinsic parameters $h_{x}$ and $h_{y}$.
Moreover, one has to adapt the camera matrix $K$ at each level of the coarse-to-fine scheme.
This requires to scale both the grid size and the principal point $(c_1,c_2)^\top$.

\section{Evaluation}
\label{sec:experimental-results}

{\bf Test Images and Error Measures.} 
In order to evaluate our novel approach we make use of four synthetic images with ground 
truth that fulfil the underlying assumptions regarding reflectance and illumination. 
This allows us to compute two error measures: one with respect to the reconstructed 
surface and the other one with respect to reprojected image. The first error measure 
is the \emph{relative surface error} (RSE) of a point wise computed Euclidean distance 
between the computed surface $\mathcal{S}$ and the ground truth surface 
$\mathcal{S}^{\mathrm{gt}}$. It is given by 
\begin{equation}
	\mathrm{RSE} = 
	\frac{ 
		\sum_{\overline{\Omega}_{\mathbf{a}}} \left| \mathcal{S}(\mathbf{x(a)}) - \mathcal{S}^{\mathrm{gt}}(\mathbf{x(a)}) \right|
		}
		{
		\sum_{\overline{\Omega}_{\mathbf{a}}} \left| \mathcal{S}^{\mathrm{gt}}(\mathbf{x(a)}) \right| 
		} \, ,
\end{equation}
where the normalisation allows to determine the reconstruction error \emph{relative} to 
the ground truth shape. This in turn makes errors of differently scaled 
surfaces comparable. The second error measure is the \emph{relative image error} 
(RIE) between the reprojected image $I$ and the given input image $I^{\mathrm{gt}}$. 
It is defined as follows 
\begin{equation}
\mathrm{RIE} = 
\frac{
	      \sum_{\overline{\Omega}_{\mathbf{a}}} \left| I(\mathbf{x(a)}) - I^{\mathrm{gt}}(\mathbf{x(a)}) \right|
	 }
	 {    
		  \sum_{\overline{\Omega}_{\mathbf{a}}} \left| I^{\mathrm{gt}}(\mathbf{x(a)}) \right|
	 } \, .
\end{equation} 
This time, however, the normalisation is performed with respect to the brightness 
of the input image to make reprojection results for input images with different 
brightness scale comparable. Summarising: While the first measure reflects how well 
the reconstruction matches the ground truth surface, the second measure determines 
how well the reprojection fits the input data.  

\begin{figure}[h!]
	\centering
	\subfigure{\label{fig:sombrero_image}
		\includegraphics[height=0.19\linewidth] 
		{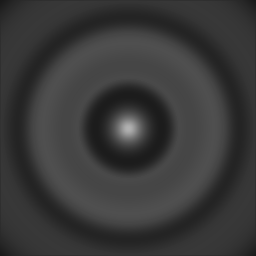}   } \hspace{-1mm}
	\subfigure{\label{fig:suzann_image}
		\includegraphics[height=0.19\linewidth]
		{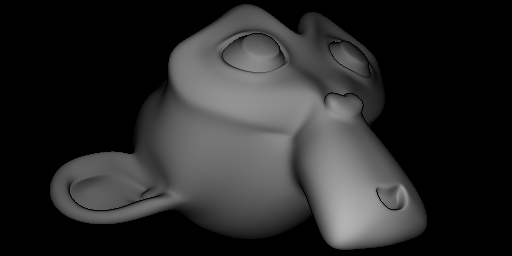}   } \hspace{-1mm}
	\subfigure{\label{fig:bunny_image}
		\includegraphics[height=0.19\linewidth]
		{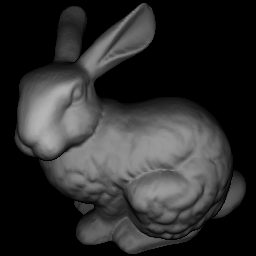}   } \hspace{-1mm}
	\subfigure{\label{fig:dragon_image}
		\includegraphics[height=0.19\linewidth] 
		{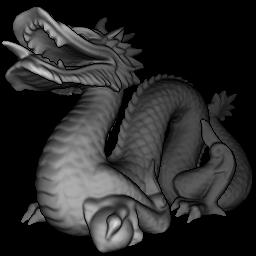}   } \\ \vspace{-2mm}

	\caption{ Synthetic test images. {\bf From left to right:} \emph{Sombrero}, \emph{Suzanne}, \emph{Stanford Bunny} and  \emph{Dragon}.  } 
	\label{fig:synthetic}
\end{figure}

Let us now discuss the considered test images which are depicted in 
Fig. \ref{fig:synthetic} in detail.
The first synthetic test image {\em Sombrero} was generated from a known 
parametric surface, using the following equation
\begin{equation}
Z(X, Y) = 0.5 \frac{\sin \left( r(X, Y) \right) }{r(X, Y)} + 1.7 \, ,
\quad
r(X, Y) = \sqrt{ \left( 10 X \right)^2  + \left( 10 Y \right)^2 } \, .
\end{equation} 
The image was rendered using Eq. \eqref{eq:15} at a size of $256\!\times\!256$ pixels, 
where the focal length was set $\mathtt{f} = 1$, the grid size was chosen to be 
$h_{x} = h_{y} = 1/200$ and the principal point was fixed at $c = (128,128)^\top$.
The second test image {\em Suzanne} was generated using the open-source software 
Blender \cite{BLENDER}. 
In this context, the Z-buffer of the rendering path and the corresponding intrinsic 
parameters ($\mathtt{f} = 35$, $h_{x}= 1/16$, $h_{y} = 9/128$, $c = (256,128)^\top$) 
were extracted and the final image was rendered at a size of $512\!\times\!256$ using 
Eq. \eqref{eq:15} as before. The other two test images {\em Stanford Bunny} and 
{\em Dragon} have been computed likewise 
using 3-D models obtained from the Stanford 3D scanning repository \cite{STAN3D}.
For them a size of $256\!\times\!256$ pixels and the intrinsic parameters 
($\mathtt{f} = 35$, $h_{x} = 1/8$, $h_{y} = 9/128$, $c = (128,128)^\top$) 
were chosen. Finally, all images were saved as $8$-bit grey-value images. 

Let us finally comment on the selection of the parameters in our experiments.
In order to keep the number of parameters low, we choose a preferred standard set of 
parameters for all the following experiments, unless otherwise stated: A downsampling 
factor of $\eta = 0.8$ for the coarse-to-fine approach, $n=10^6$ solver iterations on 
each coarse-to-fine level and a contrast parameter of $\lambda = 10^{-3}$. Moreover,
the time step size $\tau$ provided in the different experiments always refers to
the simplified explicit scheme. The time step size for the full explicit scheme is
$\min(h_{x}^{2}, h_{y}^{2})$ times smaller.

\begin{figure}[t!]
	\centering
	
	\subfigure{\label{fig:sombrero_gt}
		\includegraphics[width=0.315\linewidth] 
		{images/sombrero.png}   } \hspace{-1mm}
	\subfigure{\label{fig:sombrero_gt_depth}
		\includegraphics[width=0.315\linewidth]
		{images/bunny.png}   } \hspace{-1mm}
	\subfigure{\label{fig:sombrero_depth}
		\includegraphics[width=0.315\linewidth] 
		{images/dragon.png}   } \\ \vspace{-2mm}

	\subfigure{\label{fig:sombrero_gt-2}
		\includegraphics[width=0.315\linewidth] 
		{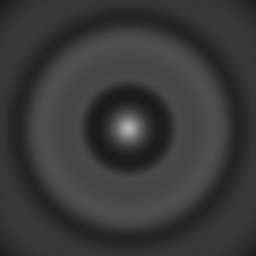}   } \hspace{-1mm}
	\subfigure{\label{fig:sombrero_gt_depth-2}
		\includegraphics[width=0.315\linewidth]
		{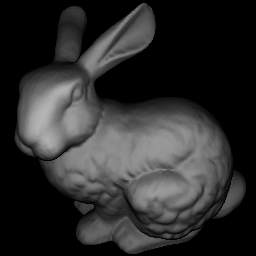}   } \hspace{-1mm}
	\subfigure{\label{fig:sombrero_depth-2}
		\includegraphics[width=0.315\linewidth] 
		{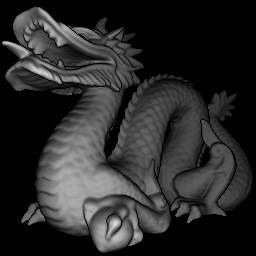}   } \\ \vspace{-2mm}	

	\subfigure{\label{fig:sombrero_gt-3}
		\includegraphics[width=0.315\linewidth] 
		{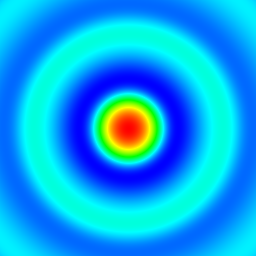}   } \hspace{-1mm}
	\subfigure{\label{fig:sombrero_gt_depth-3}
		\includegraphics[width=0.315\linewidth]
		{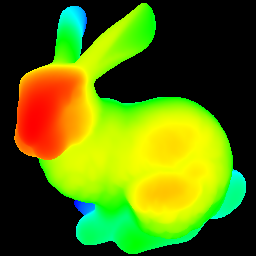}   } \hspace{-1mm}
	\subfigure{\label{fig:sombrero_depth-3}
		\includegraphics[width=0.315\linewidth] 
		{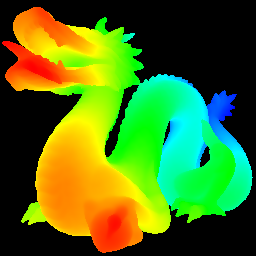}   } \\ \vspace{-2mm}

	\subfigure{\label{fig:sombrero_gt-4}
		\includegraphics[width=0.315\linewidth] 
		{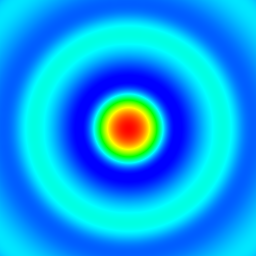}   } \hspace{-1mm}
	\subfigure{\label{fig:sombrero_gt_depth-4}
		\includegraphics[width=0.315\linewidth]
		{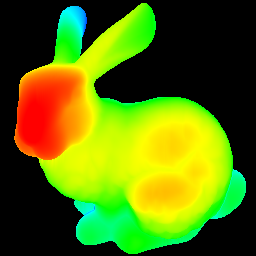}   } \hspace{-1mm}
	\subfigure{\label{fig:sombrero_depth-4}
		\includegraphics[width=0.315\linewidth] 
		{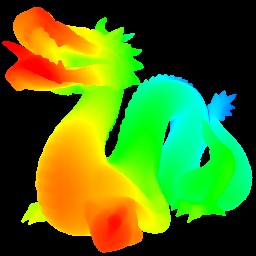}   } \\ \vspace{-2mm}	

	\caption{ {\bf First column, from top to bottom:} Input image, reprojected image, ground truth depth, computed depth for the \emph{Sombrero} test image ($\alpha = 7.5\times 10^{-5}$, $\tau = 10^{-2}$, $n=10^6$). 
		{\bf Second column:} Ditto for the \emph{Stanford Bunny} test image 
		($\alpha = 7.5\times 10^{-5}$, $\tau = 10^{-3}$, $n=10^6$). {\bf Third column:} 
		Ditto for the \emph{Dragon} test image ($\alpha = 7.5\times 10^{-8}$, $\tau = 10^{-3}$, $n=10^6$).} 
	\label{fig:synthetic-sbd}
	
	\vspace{5mm}
\end{figure}

\begin{figure}[h!]
	\centering
	
	\subfigure{\label{fig:suzanne_gt}
		\includegraphics[width=0.475\linewidth] 
		{images/suzanne.png}   } \hspace{-1mm}
	\subfigure{\label{fig:suzanne_gt_depth}
		\includegraphics[width=0.475\linewidth]
		{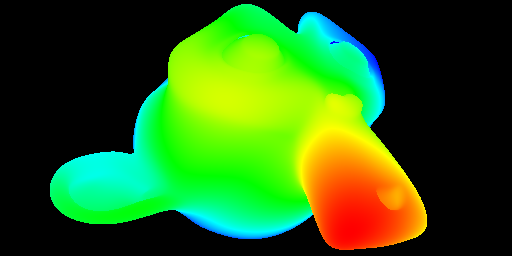}   } \\ \vspace{-2mm}
	
	\subfigure{\label{fig:suzanne}
		\includegraphics[width=0.475\linewidth] 
		{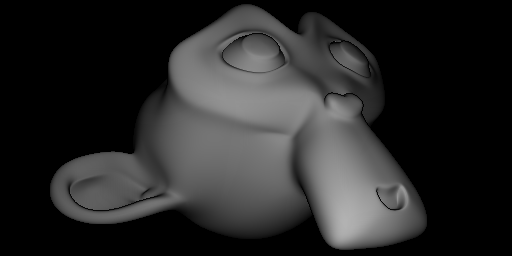}   } \hspace{-1mm}
	\subfigure{\label{fig:suzanne_depth}
		\includegraphics[width=0.475\linewidth]
		{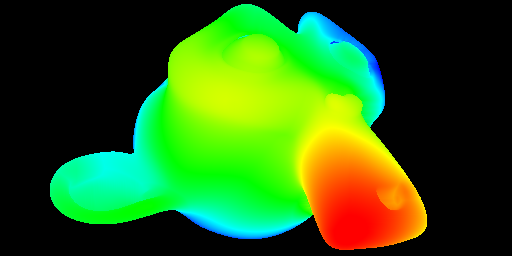}   } \\ \vspace{-2mm}
	
	\caption{{\bf First row, from left to right:} Input image and ground truth depth of the 
        \emph{Suzanne} test image. {\bf Second row:} Reprojected image and the computed depth ($\alpha = 10^{-7}$, $\tau = 10^{-3}$, $n=10^6$).}
	\label{fig:synthetic-suzanne}
	
\end{figure}

\begin{figure}[h!]
	\centering
	\subfigure{\label{fig:bunny_image-2}
		\includegraphics[height=0.235\linewidth]
		{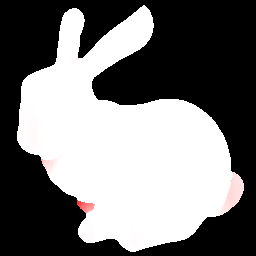}   } \hspace{-1mm}
	\subfigure{\label{fig:dragon_image-2}
		\includegraphics[height=0.235\linewidth]  
		{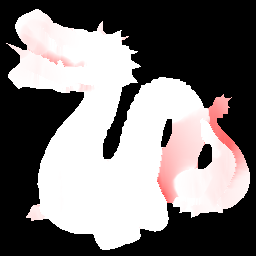}   } \hspace{-1mm}
	\subfigure{\label{fig:suzann_image-2}
		\includegraphics[height=0.235\linewidth]
		{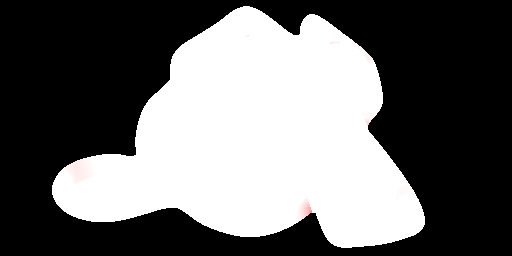}   } \\ \vspace{-2mm}
	\caption{ Surface error maps. {\bf From left to right:} \emph{Stanford Bunny}, \emph{Dragon} and \emph{Suzanne}. Red denotes errors above 1 percent, where the intensity encodes the error magnitude. White denotes errors below 1 percent. The \emph{Sombrero} is not shown, since the error is below 1 percent everywhere. } 
	\label{fig:errordiff}
\end{figure}

\noindent {\bf Results on Synthetic Test Images.} In our first experiment we evaluate the 
reconstruction quality of our novel approach. To this end, we applied our perspective SfS 
algorithm to all four of the previously discussed test images and compared the reprojected image 
and the reconstruction to the ground truth; see Fig.\ \ref{fig:synthetic-sbd} and 
Fig.\ \ref{fig:synthetic-suzanne}. Herein, the depth values are colour-coded in such a way that 
depth increases from red via green to blue. 
As one can see, both the reprojected image as well as the estimated depth values coincide very 
well with the ground truth. This is also confirmed by the corresponding surface error maps in Fig.\ \ref{fig:errordiff}. Indeed, only small differences for the \emph{Stanford bunny} (right paw) 
and the \emph{Dragon} (tail tip) are visible. As a consequence both error measures which 
are listed in Table \ref{tab:resultsSyntheticQuadtratic} are very small. Moreover, one 
can see that the proposed
subquadratic penaliser outperforms a quadratic smoothness term in most cases. Only for the {\em Sombrero}
which has a rather smooth surface, the reconstruction error is smaller in the quadratic case.\\

\begin{table}[h!]
	\caption{Results for our approach with quadratic and subquadratic penaliser. Error measures 
		are given in terms of the relative surface error (RSE) and the relative image error (RIE). 
		Best results for each test image are highlighted boldface. Same parameters as in Fig.\ \ref{fig:synthetic-sbd} and 
		Fig.\ \ref{fig:synthetic-suzanne}.}
	\label{tab:resultsSyntheticQuadtratic}
	\begin{center}
		\begin{tabular}{|clc||ccc|ccc||ccc|ccc||ccc||} \hline 
			\multicolumn{3}{|c||}{} && \multicolumn{4}{c}{} &&&  \multicolumn{4}{c}{}  &&& &\\[-2mm]
			\multicolumn{3}{c|}{} && \multicolumn{4}{c}{quadratic} &&& \multicolumn{4}{c}{subquadratic} &&&  &\\[-2mm]
			\multicolumn{3}{|c||}{} && \multicolumn{4}{c}{} &&&  \multicolumn{4}{c}{}  &&& &\\ \cline{4-15}
			\multicolumn{3}{|c||}{} && &&&  &&&  &&&  &&& &\\[-2mm]
			\multicolumn{3}{|c||}{} && RSE &&& RIE &&& RSE &&& RIE &&& runtime &\\[-2mm]
			\multicolumn{3}{|c||}{} && &&&  &&&  &&& &&&  &\\\hline \hline
			& &&& &&&  &&&  &&&  &&& &\\[-2mm]
			& Sombrero &&& $\textbf{0.00208}$ &&& $0.00694$ &&& $0.00318$ &&& $\textbf{0.00209}$ &&&
			$29113$s 
			&\\[-2mm]
			& &&& &&&  &&&  &&& &&& &\\\hline
			& &&& &&&  &&&  &&&  &&& &\\[-2mm]
			& Stanford Bunny &&& $0.00546$ &&& $0.00015$ &&& $\textbf{0.00439}$ &&& $\textbf{0.00007}$ &&&
			$23969$s 
			&\\[-2mm]
			& &&& &&&  &&&  &&& &&& &\\\hline
			& &&& &&&  &&&  &&& &&& &\\[-2mm]
			& Dragon &&& $\textbf{0.01376}$ &&& $\textbf{0.00028}$ &&& $\textbf{0.01376}$ &&& $\textbf{0.00028}$ &&& 
			$25350$s 
			&\\[-2mm]
			& &&& &&&  &&&  &&& &&& &\\\hline
			& &&& &&&  &&&  &&&   &&&  &\\[-2mm]
			& Suzanne &&& $0.00392$ &&& $0.00011$ &&& $\textbf{0.00251}$ &&& $\textbf{0.00002}$ &&& 
			$48395$s 
			&\\[-2mm]
			& &&& &&&  &&&  &&& &&& &\\\hline 
		\end{tabular}
	\end{center}
\end{table}

\begin{figure}[hbt]	
	\centering
	\includegraphics[width=0.5\linewidth]{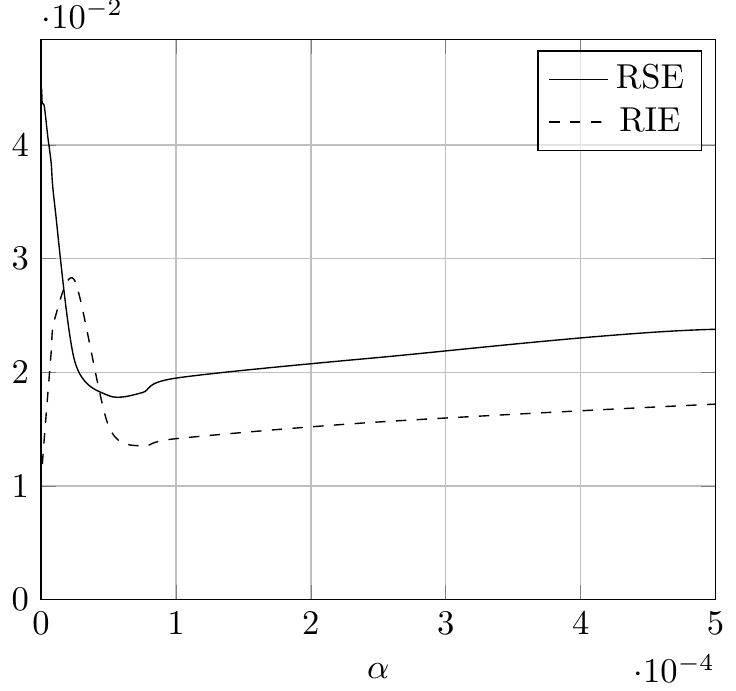} 
	\caption{Impact of the amount of regularisation on the reconstruction quality
        and the reprojection accuracy for the \emph{Sombrero} test image.}
	\label{fig:InfluenceRegularization}
\end{figure}

\noindent {\bf Influence of the Regularisation.} In our second experiment we investigate the 
influence of the regularisation on the quality of the reconstruction and its reprojection. 
To this end, we consider the \emph{Sombrero} test image and vary the regularisation parameter 
$\alpha$ while the other parameters are kept fixed ($\tau = 0.001$, $n=10^4$). 
The outcome is visualised in Fig.\ \ref{fig:InfluenceRegularization}. While the reprojection 
related error measure (RIE) increases for a moderate amount of regularisation but
is overall very low, the surface related error measure (RSE) decreases by almost a 
factor three (from $4.4 \times 10^{-2}$ to $1.7 \times 10^{-2}$). This, however, is not 
surprising, since the computed surface typically exhibits some form of smoothness and thus 
benefits from a moderate amount of regularisation. Since the actual purpose of SfS is 
to find the correct surface, this shows that the regularisation may have an overall positive 
impact on the quality of the results. \\

\noindent {\bf Independence of the Initialisation.} In our third experiment we analyse the 
dependency of our approach on the initialisation. To this end, we use the \emph{Stanford Bunny} ($z \in [1,2]$) and compare our initialisation on the coarsest scale of the proposed 
coarse-to-fine scheme (cf. Eq. \eqref{eq:333}) with two other initialisations based on plain 
surfaces ($z=1$, $z=10$). The initial error and the outcome after $n=10^6$ iterations are 
listed in Table \ref{tab:resultsInit}. While the initial error for a good guess ($z=1$) and 
a poor initialisation ($z=10$) differs significantly, the quality of the reconstruction and 
the reprojection is identical after sufficiently many iterations. This also holds for our 
initialisation which can be computed from the input image without requiring a specific 
knowledge of the depth. That all initialisations converge to the same solution, however, is 
not surprising since the estimation is embedded in our coarse-to-fine scheme. \\

\begin{table}[t!]
	\caption{Impact of different initialisations on the reconstruction quality and 
        reprojection accuracy for the \emph{Stanford Bunny} 
        ($\alpha = 7.5 \times 10^{-5}$, $\tau = 10^{-3}$, $n=10^6$).}
	\label{tab:resultsInit}
	\begin{center}
		\begin{tabular}{|clc||ccc|ccc||ccc|ccc|} \hline
			\multicolumn{3}{|c||}{} && \multicolumn{4}{c}{} &&&  \multicolumn{4}{c}{} &\\[-2mm]
			\multicolumn{3}{|c||}{} && \multicolumn{4}{c}{initial error} &&& \multicolumn{4}{c}{after computation} &\\[-2mm]
			\multicolumn{3}{|c||}{} && \multicolumn{4}{c}{} &&&  \multicolumn{4}{c}{} &\\ \cline{4-15}
			\multicolumn{3}{|c||}{} && &&&  &&&  &&&  &\\[-2mm]
			\multicolumn{3}{|c||}{} && RSE &&& RIE &&& RSE &&& RIE &\\[-2mm]
			\multicolumn{3}{|c||}{} && &&&  &&&  &&&  &\\\hline \hline
			& &&& &&&  &&&  &&&   &\\[-2mm]
			& plane ($z = 1$)  &&& $0.25804$ &&& $1.63174$ &&& $0.00439$ &&& $0.00007$ &\\[-2mm]
			& &&& &&&  &&&  &&&   &\\\hline
			& &&& &&&  &&&  &&&   &\\[-2mm]
			& plane ($z = 10$) &&& $6.41960$ &&& $0.97373$ &&& $0.00439$ &&& $0.00007$ &\\[-2mm]
			& &&& &&&  &&&  &&&   &\\\hline
			& &&& &&&  &&&  &&&   &\\[-2mm]
			& proposed &&& $0.37712$ &&& $0.74363$ &&& $0.00439$ &&& $0.00007$ &\\[-2mm]
			& &&& &&&  &&&  &&&   &\\\hline
		\end{tabular}
	\end{center}
\end{table}

\noindent {\bf Comparison of Numerical Schemes.} In our fourth experiment we compare the different 
numerical schemes proposed in Section \ref{sec:minimisation}: the full explicit scheme, the simplified 
explicit scheme and the alternating explicit scheme. In the first part of the experiment we juxtapose 
the quality of the different numerical schemes for equal stopping times (iterations $\times$ time 
step size). As one can see from the results in Table \ref{tab:resultsDirectionSmallChar}, the full explicit 
scheme clearly gives the best results in terms of reconstruction quality and reprojection accuracy. 
However, this comes at the expense of a significantly larger runtime, since more iterations are 
needed due to the time step restrictions discussed in Section \ref{sec:minimisation}. In fact the 
runtime is up to four orders of magnitude larger making the approach hardly feasible for larger image 
sizes. In the second part of the experiment we compared the numerical schemes for an equal number of 
iterations. From the results in Table \ref{tab:resultsSmallct} it becomes evident that in this case 
the simplified explicit scheme and in particular the alternating explicit scheme perform best in most cases in terms of reconstruction quality and reprojection accuracy. This demonstrates that it can be worthwhile to (partly) omit 
the terms that are added in the full explicit scheme
since they slow down 
the convergence, but doing so does not necessarily compromise the quality. 
\begin{table}[h!]
	\caption{Comparison of different numerical schemes for equal stopping time 
        $t = n \times \tau$. Results and runtimes refer to smaller versions of the four test 
        images. Same parameters as in Fig.\ \ref{fig:synthetic-sbd} and 
        Fig.\ \ref{fig:synthetic-suzanne} except for $n$, which is given by $n = t / \tau$. }
	\label{tab:resultsDirectionSmallChar}
	\begin{center}
		\begin{tabular}{|clc||cccccc||cccccc||cccccc|} \hline
			\multicolumn{3}{|c||}{} && \multicolumn{4}{c}{} &&& \multicolumn{4}{c}{} &&& \multicolumn{4}{c}{}  &\\[-2mm]
			\multicolumn{3}{|c||}{} && \multicolumn{4}{c}{alternating scheme} &&& \multicolumn{4}{c}{simplified scheme} &&& \multicolumn{4}{c}{full scheme} &\\[-2mm]
			\multicolumn{3}{|c||}{} && \multicolumn{4}{c}{} &&& \multicolumn{4}{c}{} &&&  \multicolumn{4}{c}{}  &\\ \hline
			&  &&&  &&&  &&&  &&&  &&&  &&& &\\[-2mm]
			& test image  &&& RSE  &&& RIE &&& RSE  &&& RIE &&& RSE &&& RIE &\\[-2mm]
			&  &&&  &&&  &&&  &&&  &&&  &&&  &\\\hline \hline
			& &&& &&&  &&&  &&&  &&&  &&& &\\[-2mm]
			& Small Sombrero  &&& $0.01823$ &&& $0.01920$ &&& $0.01820$ &&& $0.02048$ &&& $\textbf{0.00785}$ &&& $\textbf{0.00527}$ &\\[-2mm] 
			& &&& \multicolumn{4}{c}{} &&& \multicolumn{4}{c}{} &&& \multicolumn{4}{c}{} &\\[-1mm]
			& ($128 \times 128$) &&&  \multicolumn{4}{c}{(runtime: $30$s)}  &&& \multicolumn{4}{c}{(runtime: $15$s)} &&&  \multicolumn{4}{c}{(runtime: $178021$s)} &\\[-3mm]
			& &&& \multicolumn{4}{c}{}  &&&  \multicolumn{4}{c}{}  &&&  \multicolumn{4}{c}{} &\\\hline
			& &&& \multicolumn{4}{c}{} &&& \multicolumn{4}{c}{} &&& \multicolumn{4}{c}{} &\\[-2mm]
			& Small Stanford Bunny &&& $0.00659$ &&& $0.00151$ &&& $0.00667$ &&& $0.00257$ &&& $\textbf{0.00576}$ &&& $\textbf{0.00097}$ &\\[-2mm]
			& &&& &&&  &&&  &&&  &&&  &&& &\\[-1mm]
			& ($128 \times 128$) &&& \multicolumn{4}{c}{(runtime: $303$s)} &&&  \multicolumn{4}{c}{(runtime: $150$s)} &&&  \multicolumn{4}{c}{(runtime: $4278$s)} &\\[-3mm]
			& &&& \multicolumn{4}{c}{}  &&&  \multicolumn{4}{c}{}  &&&  \multicolumn{4}{c}{} &\\\hline
			& &&& \multicolumn{4}{c}{} &&& \multicolumn{4}{c}{} &&& \multicolumn{4}{c}{} &\\[-2mm]
			& Small Dragon &&& $0.01667$ &&& $0.00267$ &&& $0.01673$ &&& $0.00620$ &&& $\textbf{0.01526}$ &&& $\textbf{0.00205}$ &\\[-2mm]
			& &&& &&&  &&&  &&&  &&&  &&& &\\[-1mm]
			& ($128 \times 128$) &&& \multicolumn{4}{c}{(runtime: $308$s)} &&& \multicolumn{4}{c}{(runtime: $149$s)} &&&  \multicolumn{4}{c}{(runtime: $4304$s)} &\\[-3mm]
			& &&& \multicolumn{4}{c}{}  &&&  \multicolumn{4}{c}{}  &&&  \multicolumn{4}{c}{} &\\\hline
			& &&& \multicolumn{4}{c}{} &&& \multicolumn{4}{c}{} &&& \multicolumn{4}{c}{} &\\[-2mm]
			& Small Suzanne &&& $\textbf{0.00899}$ &&& $0.00514$ &&& $0.01055$ &&& $0.01909$ &&& $0.01022$ &&& $\textbf{0.00203}$ &\\[-2mm]
			& &&& &&&  &&&  &&&  &&&  &&& &\\[-1mm]
			& ($128 \times 96$) &&& \multicolumn{4}{c}{(runtime: $223$s)} &&&  \multicolumn{4}{c}{(runtime: $111$s)} &&& \multicolumn{4}{c}{(runtime: $2384$s)} &\\[-3mm]
			& &&& \multicolumn{4}{c}{}  &&&  \multicolumn{4}{c}{}  &&&  \multicolumn{4}{c}{} &\\\hline
		\end{tabular}
	\end{center}
\end{table}

\begin{table}[h!]
	\caption{Comparison of different numerical schemes for equal number of iterations. Results 
        refer to the smaller versions of the four test test images, see Tab.\ \ref{tab:resultsDirectionSmallChar}. 
        The same parameters as in Fig.\ \ref{fig:synthetic-sbd} and Fig.\ \ref{fig:synthetic-suzanne} have been used except for $n$, which is given by $n = 10^7$. }
	\label{tab:resultsSmallct}
	\begin{center}
		\begin{tabular}{|clc||ccc|ccc||ccc|ccc||ccc|ccc|} \hline
			\multicolumn{3}{|c||}{} && \multicolumn{4}{c}{} &&& \multicolumn{4}{c}{} &&& \multicolumn{4}{c}{}  &\\[-2mm]
			\multicolumn{3}{|c||}{} && \multicolumn{4}{c}{alternating scheme} &&& \multicolumn{4}{c}{simplified scheme} &&& \multicolumn{4}{c}{full scheme} &\\[-2mm]
			\multicolumn{3}{|c||}{} && \multicolumn{4}{c}{} &&& \multicolumn{4}{c}{} &&&  \multicolumn{4}{c}{}  &\\ \hline 
			&  &&&  &&&  &&&  &&&  &&&  &&& &\\[-2mm]
			& test image &&& 
			RSE &&& RIE &&& RSE &&& RIE &&& RSE &&& RIE &\\[-2mm]
			&  &&&  &&&  &&&  &&&  &&&  &&&  &\\\hline \hline
			& &&& &&&  &&&  &&&  &&&  &&& &\\[-2mm]
			& Small Sombrero &&& $0.02357$ &&& $\textbf{0.00082}$ &&& $0.02392$ &&& $0.00659$ &&& $\textbf{0.00358}$ &&& $0.00319$ &\\[-2mm]
			& &&& &&&  &&&  &&&  &&&  &&& &\\\hline
			& &&& &&&  &&&  &&&  &&&  &&& &\\[-2mm]
			& Small Stanford Bunny &&& $0.00390$ &&& $\textbf{0.00001}$ &&& $\textbf{0.00378}$ &&& $0.00004$ &&& $0.00489$ &&& $0.00047$ &\\[-2mm]
			& &&& &&&  &&&  &&&  &&&  &&& &\\\hline
			& &&& &&&  &&&  &&&  &&&  &&& &\\[-2mm]
			& Small Dragon &&& $0.00572$ &&& $\textbf{0.00001}$ &&& $\textbf{0.00562}$ &&& $\textbf{0.00001}$ &&& $0.00964$ &&& $0.00170$ &\\[-2mm]
			& &&& &&&  &&&  &&&  &&&  &&& &\\\hline
			& &&& &&&  &&&  &&&  &&&  &&& &\\[-2mm]
			& Small Suzanne &&& $\textbf{0.00319}$ &&& $0.00002$ &&& $0.00320$ &&& $\textbf{0.00001}$ &&& $0.00505$ &&& $0.00056$ &\\[-2mm]
			& &&& &&&  &&&  &&&  &&&  &&& &\\\hline
		\end{tabular}
	\end{center}
	\vspace{3mm}
\end{table}

\begin{table}[h!]
	\caption{Evaluation of inpainting properties for degraded versions of the \emph{Stanford Bunny} test image. Same parameters as in Figure \ref{fig:synthetic-inpaint}.}
	\label{tab:resultsInpaint}
	\begin{center}
		\begin{tabular}{|clc||ccc|ccc|ccc|} \hline
			\multicolumn{3}{|c||}{} && &&& &&& &\\[-2mm]
			\multicolumn{3}{|c||}{} && perforated version &&& sliced version &&& original version &\\[-2mm]
			\multicolumn{3}{|c||}{} && &&& &&& &\\[-2mm]
			\multicolumn{3}{|c||}{} && (Fig.\ \ref{fig:synthetic-inpaint}, top row) &&& (Fig.\ \ref{fig:synthetic-inpaint}, bottom row) &&& (Fig.\ \ref{fig:synthetic}) &\\[-2mm]
			\multicolumn{3}{|c||}{} && &&& &&& &\\\hline \hline
			& &&& &&& &&& &\\[-2mm]
			& RSE &&& $0.00439$ &&& $0.00509$ &&& $0.00439$ &\\[-2mm]
			& &&& &&& &&& &\\\hline
			& &&& &&& &&& &\\[-2mm]
			& RIE &&& $0.00039$ &&& $0.00249$ &&& $0.00007$ &\\[-2mm]
			& &&& &&& &&& &\\\hline
		\end{tabular}
	\end{center}
	\vspace{3mm}
\end{table}

\begin{figure}[t!]
	\centering
	
	\subfigure{\label{fig:bunny_per_inpaint_gt_}
		\includegraphics[width=0.23\linewidth] 
		{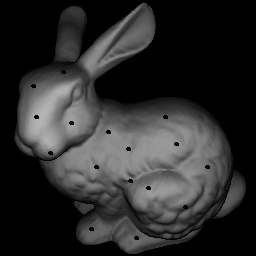}   } \hspace{-1mm}
	\subfigure{\label{fig:bunny_per_inpaint_mask}
		\includegraphics[width=0.23\linewidth]
		{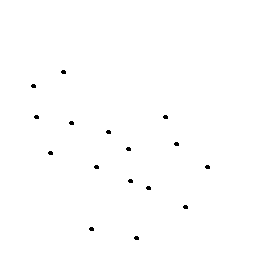}   } \hspace{-1mm}
	\subfigure{\label{fig:bunny_per_depth}
		\includegraphics[width=0.23\linewidth] 
		{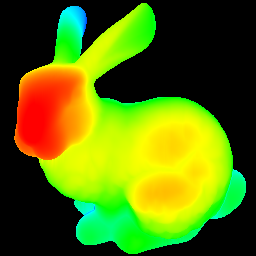}   } \hspace{-1mm}
	\subfigure{\label{fig:bunny_per}
		\includegraphics[width=0.23\linewidth]
		{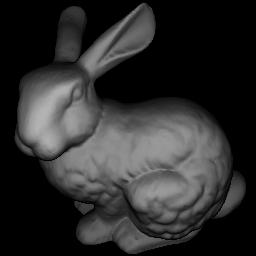}   } \\ \vspace{-2mm}
	
	\subfigure{\label{fig:bunny_sh_inpaint_gt}
		\includegraphics[width=0.23\linewidth] 
		{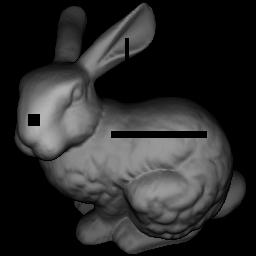}   } 
	\hspace{-1mm}
	\subfigure{\label{fig:bunny_sh_inpaint_mask}
		\includegraphics[width=0.23\linewidth]
		{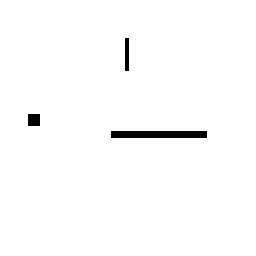}   } \hspace{-1mm}
	\subfigure{\label{fig:bunny_sh_depth}
		\includegraphics[width=0.23\linewidth]
		{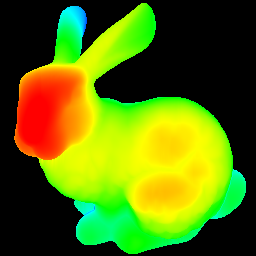} } \hspace{-1mm}
	\subfigure{\label{fig:bunny_sh}
		\includegraphics[width=0.23\linewidth]
		{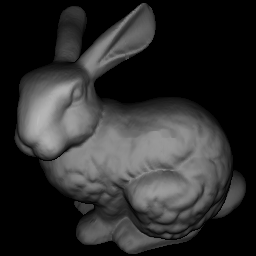} } \\ \vspace{-2mm}

	\caption{ {\bf First row, from left to right:} Perforated version of the \emph{Stanford Bunny} test image, corresponding confidence function $c$, computed depth values,  reprojected image ($\alpha = 7.5\times 10^{-5}$, $\tau = 10^{-3}$, $n = 10^6$). {\bf Second row:} Ditto for the sliced version (same parameters).} 
	\label{fig:synthetic-inpaint}
	
\end{figure}

\noindent {\bf Reconstruction with Inpainting.} In our fifth experiment we demonstrate 
the inpainting capabilities of the regularisation in combination with the confidence 
function $c$ embedded in the data term. For this reason we created a pair of degraded 
\emph{Stanford Bunny} test images together with the corresponding confidence functions, 
which are both depicted in Fig.\ \ref{fig:synthetic-inpaint}. In addition, the computed 
depth values and the reprojected images are shown. One can see that in both cases the 
missing regions in the input image can hardly deteriorate the quality of the results 
since the smoothness term fills in the information from the neighbourhood. This is also reflected in the error measures given in Table \ref{tab:resultsInpaint}.
In case of the perforated version the surface error even remains the same compared to the 
result for the original version.\\

\begin{figure}[t!]
	\centering

	\subfigure{\label{fig:bunny_noise_gt}
		\includegraphics[width=0.23\linewidth,height=0.23\linewidth]  
		{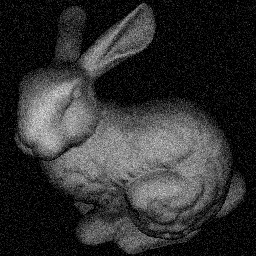}   } \hspace{-1mm}
	\subfigure{\label{fig:bunny_noise_gt_depth}
		\includegraphics[width=0.23\linewidth,height=0.23\linewidth]
		{images/bunny-depth.png}   } \hspace{-1mm}
	\subfigure{\label{fig:bunny_depth}
		\includegraphics[width=0.23\linewidth,height=0.23\linewidth] 
		{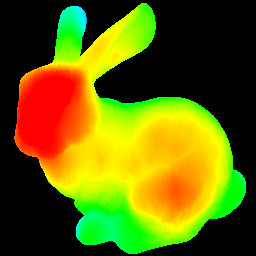}   } \hspace{-1mm}
	\subfigure{\label{fig:bunny}
		\includegraphics[width=0.23\linewidth,height=0.23\linewidth]
		{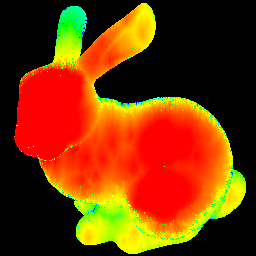}   } \\ \vspace{-2mm}
	
	\caption{ {\bf From left to right:} Noisy version of the \emph{Stanford Bunny} (Gaussian noise with $\sigma=20$), ground truth depth, computed depth using our variational approach ($\alpha = 1.0$, $\tau = 10^{-5}$, $n=10^6$), 
		computed depth using the PDE-based approach of Vogel {\em et al.} \cite{Vogel_SSVM2009} with Lambertian  model.} 
	\label{fig:bunnyNoise}
\end{figure}

\noindent {\bf Comparison with a PDE-Based Approach.} In our seventh experiment we compare the results of our variational method with the PDE-based approach of Vogel {\em et al.} \cite{Vogel_SSVM2009} with Lambertian reflectance model. This essentially comes down to a comparison to the baseline method of Prados {\em et al.} \cite{Prados_CVPR2005} which is solved by Vogel {\em et al.} \cite{Vogel_SSVM2009} as part of a Phong-based model using an efficient fast marching scheme \cite{Sethian1999}. In this experiment we consider two scenarios, that nicely demonstrate the advantages and shortcomings of the different types of methods: On the one hand, we use input images without noise, on the other hand, we added Gaussian noise of standard deviation $\sigma=20$ before applying the two methods. The corresponding results are summarised in Tables \ref{tab:comparison} and \ref{tab:comparisonNoise}, respectively. For the test images without noise both approaches give excellent results with errors among or below $1$ 
percent of the solution. Thereby the approach of Vogel {\em et al.} gives slightly better results in terms of the relative surface error (RSE), while the variational approach gives better results in terms of the relative image error (RIE). From the viewpoint of the variational approach this can be explained as follows: While the data term penalises deviations from the photometric reprojection error and thus gives rather small RIE values, the regulariser and the coarse-to-fine scheme yield a moderate smoothing of the surface resulting in slightly higher RSE values. In the case of the noisy input images the findings are completely different. Here, the variational method can take advantage of both the regulariser and the independence of the initialisation. While a higher smoothness weight allows to obtain a smooth surface, the hierarchical initialisation via the coarse-to-fine scheme does not require to rely on noisy solutions at critical points as the PDE-based approach of Vogel {\em et al.} As a consequence, 
the resulting surface errors of $3$ to $6$ percent for our variational approach are significantly lower than those of the PDE-based model ($11$ to $20$ percent). This can also be seen from the depth estimates for the Stanford Bunny depicted in Figure \ref{fig:bunnyNoise}. Not surprisingly our findings are in full accordance with the observation in \cite{JBB2014}, in which the robustness of variational methods for perspective SfS has been investigated.\\

\begin{table}[h!]
	\caption{Comparison between our variational method and the PDE-based approach of Vogel {\em et al.} \cite{Vogel_SSVM2009} with Lambertian reflectance model ($=$ baseline model of Prados {\em et al.} \cite{Prados_CVPR2005}). Error measures 
		are given in terms of the relative surface error (RSE) and the relative image error (RIE). Same parameters as in Fig.\ \ref{fig:synthetic-sbd} and Fig.\ \ref{fig:synthetic-suzanne}.}
	\label{tab:comparison}
	\begin{center}
		\begin{tabular}{|clc||ccc|ccc||ccc|ccc|} \hline 
			\multicolumn{3}{|c||}{} &&  \multicolumn{4}{c}{} &&&  \multicolumn{4}{c}{} &\\[-2mm]
			\multicolumn{3}{c|}{} && \multicolumn{4}{c}{Vogel {\em et al.} \cite{Vogel_SSVM2009}} &&& \multicolumn{4}{c}{our method} &\\[-2mm]
			\multicolumn{3}{|c||}{} &&  \multicolumn{4}{c}{} &&& \multicolumn{4}{c}{}  &\\[-1.5mm]
			\multicolumn{3}{c|}{} && \multicolumn{4}{c}{(PDE-based approach)} &&& \multicolumn{4}{c}{(variational method)} &\\[-2mm]
			\multicolumn{3}{|c||}{} &&  \multicolumn{4}{c}{} &&& \multicolumn{4}{c}{}  &\\ \cline{4-15}
			\multicolumn{3}{|c||}{} &&     &&&     &&&     &&&     &\\[-2mm]
			\multicolumn{3}{|c||}{} && RSE &&& RIE &&& RSE &&& RIE &\\[-2mm]
			\multicolumn{3}{|c||}{} &&     &&&     &&&     &&&     &\\\hline \hline
			& &&&  &&&  &&&  &&&  &\\[-2mm]
			& Sombrero &&&  $0.00301$ &&& $0.00495$ &&& $0.00318$ &&& $0.00209$ &\\[-2mm]
			& &&&  &&&  &&&  &&&  &\\\hline
			& &&&  &&&  &&&  &&&  &\\[-2mm]
			& Stanford Bunny &&&  $0.00266$ &&& $0.00154$  &&& $0.00439$ &&& $0.00007$ &\\[-2mm]
			& &&&  &&&  &&&  &&&  &\\\hline
			& &&&  &&&  &&&  &&&  &\\[-2mm]
			& Dragon &&&  $0.00422$ &&& $0.00255$  &&& $0.01376$ &&& $0.00028$ &\\[-2mm]
			& &&&  &&&  &&&  &&&  &\\\hline
			& &&&  &&&  &&&  &&&  &\\[-2mm]
			& Suzanne &&& $0.00253$ &&& $0.00082$  &&& $0.00251$ &&& $0.00002$ &\\[-2mm]
			& &&&  &&&  &&&  &&&  &\\\hline  
		\end{tabular}
	\end{center}
\end{table}

\begin{table}[h!]
	\caption{Performance under noise. Comparison between our variational method and the PDE-based approach of Vogel {\em et al.} \cite{Vogel_SSVM2009} with Lambertian reflectance model ($=$ baseline model of Prados {\em et al.} \cite{Prados_CVPR2005}). Gaussian noise of standard deviation $\sigma=20$. Error measures 
		are given in terms of the relative surface error (RSE) and the relative image error (RIE). The applied parameters are as follows: \emph{Sombrero} ($\alpha = 0.1$, $\tau = 10^{-5}$, $n=10^6$), \emph{Stanford Bunny} ($\alpha = 1.0$, $\tau = 10^{-5}$, $n=10^6$), \emph{Dragon} ($\alpha = 1.0$, $\tau = 10^{-5}$, $n=10^6$), \emph{Suzanne} ($\alpha = 1.0$, $\tau = 5\times10^{-6}$, $n=10^6$).}
	\label{tab:comparisonNoise}
	\begin{center}
		\begin{tabular}{|clc||ccc|ccc||ccc|ccc|} \hline  
			\multicolumn{3}{|c||}{} &&  \multicolumn{4}{c}{} &&&  \multicolumn{4}{c}{} &\\[-2mm]
			\multicolumn{3}{c|}{} && \multicolumn{4}{c}{Vogel {\em et al.} \cite{Vogel_SSVM2009}} &&& \multicolumn{4}{c}{Our method.} &\\[-2mm]
			\multicolumn{3}{|c||}{} &&  \multicolumn{4}{c}{} &&& \multicolumn{4}{c}{}  &\\[-1.5mm]
			\multicolumn{3}{c|}{} && \multicolumn{4}{c}{(PDE-based approach)} &&& \multicolumn{4}{c}{(variational method)} &\\[-2mm]
			\multicolumn{3}{|c||}{} &&  \multicolumn{4}{c}{} &&& \multicolumn{4}{c}{}  &\\ \cline{4-15}
			\multicolumn{3}{|c||}{} &&     &&&     &&&     &&&     &\\[-2mm]
			\multicolumn{3}{|c||}{} && RSE &&& RIE &&& RSE &&& RIE &\\[-2mm]
			\multicolumn{3}{|c||}{} &&     &&&     &&&     &&&     &\\\hline \hline
			& &&&  &&&  &&&  &&&  &\\[-2mm]
			& Noisy Sombrero &&&  $0.19530$ &&& $0.27254$ &&& $0.05118$ &&& $0.13239$ &\\[-2mm]
			& &&&  &&&  &&&  &&&  &\\\hline
			& &&&  &&&  &&&  &&&  &\\[-2mm]
			& Noisy Stanford Bunny &&&  $0.10973$ &&& $0.17347$  &&& $0.03235$ &&& $0.15279$ &\\[-2mm]
			& &&&  &&&  &&&  &&&  &\\\hline
			& &&&  &&&  &&&  &&&  &\\[-2mm]
			& Noisy Dragon &&&  $0.12240$ &&& $0.19409$  &&& $0.05395$ &&& $0.18767$ &\\[-2mm]
			& &&&  &&&  &&&  &&&  &\\\hline
			& &&&  &&&  &&&  &&&  &\\[-2mm]
			& Noisy Suzanne &&& $0.12134$ &&& $0.16783$  &&& $0.01256$ &&& $0.14302$ &\\[-2mm]
			& &&&  &&&  &&&  &&&  &\\\hline 
		\end{tabular}
	\end{center}
\end{table}

\begin{figure}[t!]
	\centering
	\subfigure{\label{fig:prados1}
		\includegraphics[width=0.23\linewidth]
		{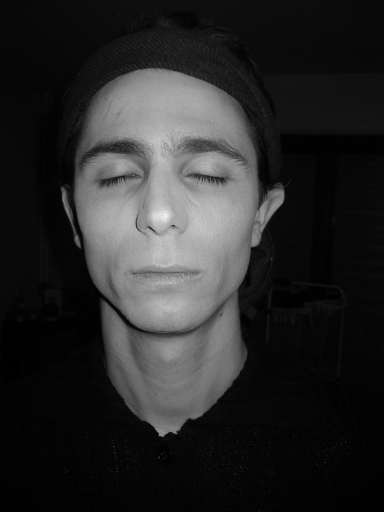}   } \hspace{-1mm}
	\subfigure{\label{fig:prados1_view1}
		\includegraphics[width=0.23\linewidth]
		{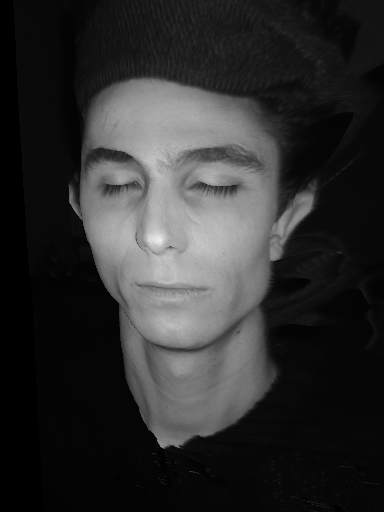}   } \hspace{-1mm}
	\subfigure{\label{fig:prados1_view2}
		\includegraphics[width=0.23\linewidth] 
		{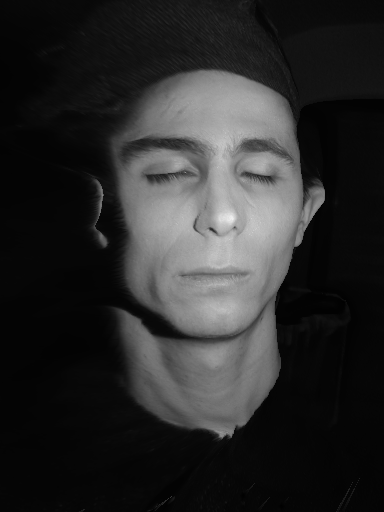}   } \hspace{-1mm}
	\subfigure{\label{fig:prados1_view3}
		\includegraphics[width=0.23\linewidth]
		{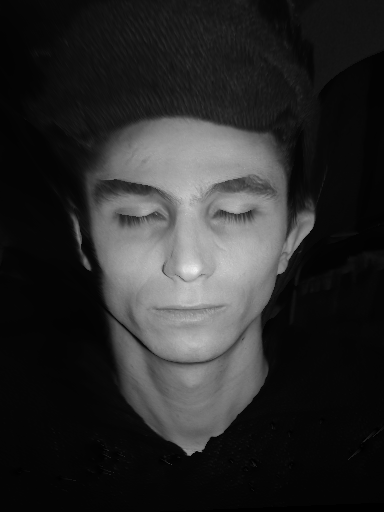}   } \\ \vspace{-2mm}
	\subfigure{\label{fig:prados2}
		\includegraphics[width=0.23\linewidth]
		{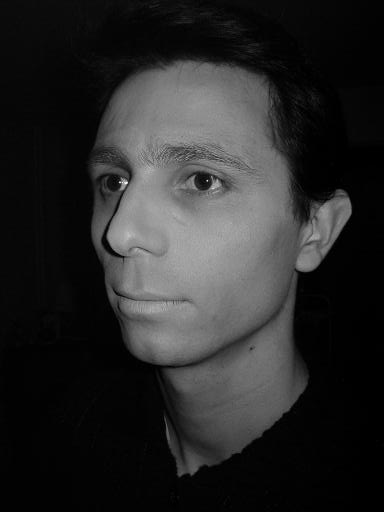}   } \hspace{-1mm}
	\subfigure{\label{fig:prados2_view1}
		\includegraphics[width=0.23\linewidth]
		{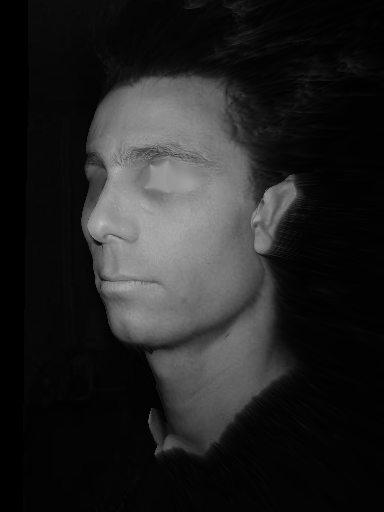}   } \hspace{-1mm}
	\subfigure{\label{fig:prados2_view2}
		\includegraphics[width=0.23\linewidth]
		{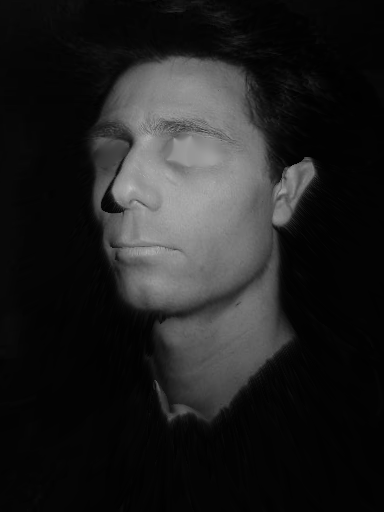}   } \hspace{-1mm}
	\subfigure{\label{fig:prados2_view3}
		\includegraphics[width=0.23\linewidth]
		{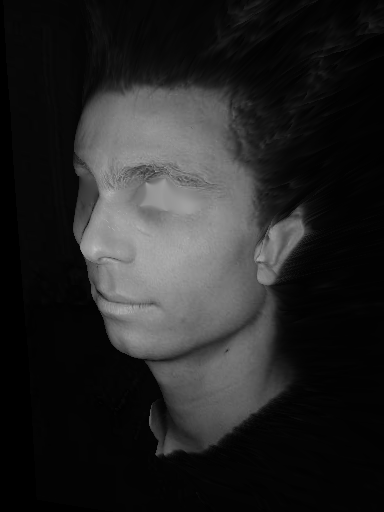}   } \\ \vspace{-2mm}
	\caption{ {\bf First row, from left to right:} Face with closed eyes, three images rendered from a new viewpoint using the estimated depth ($\alpha = 7.5 \times 10^{-5}$, $\tau = 5 \times 10^{-3}$, $n=2 \times 10^5$). {\bf Second row:} Ditto for the second test image ($\alpha = 7.5 \times 10^{-5}$, $\tau = 5 \times 10^{-3}$, $n=2 \times 10^5$).} 
	\label{fig:realworld}
	
\end{figure}

\noindent {\bf Results on Real-World Images.} Finally, in order to evaluate our approach on real-world images, we used two images of faces provided by Prados \cite{PCF06}. 
According to Prados, these images have been taken with a cheap digital camera in a dark place, where the scene is illuminated by the flash of the camera. 
The focal length is $\mathtt{f}=5.8$mm and the grid size is approximately 
$h_{x} = h_{y} = 0.018$mm. The test images as well as additional images rendered 
from a new viewpoint using the computed depth are shown in Fig.\ \ref{fig:realworld}. 
In both cases the results look quite realistic. One can also see how the depth values 
at the eyes have been inpainted in the reconstruction, since a manually defined 
confidence function was used to mask out those regions where the assumption of 
a Lambertian surface is violated. \\

\section{Conclusion}
\label{sec:conclusion}

In this paper, we described a novel variational model for perspective shape from
shading that not only has many desirable theoretical properties but also yields
very convincing reconstruction results for synthetic and real-world input images,
even in the presence of noise or other deteriorations in an input image.
While the arising optimisation problem has turned out to be challenging, we have proposed
an alternating explicit scheme embedded in a coarse-to-fine framework that is robust with respect 
to the initialisation and that allows reasonable computation times compared to a standard explicit scheme.  

Besides the results that are documented via extensive experiments in this chapter, 
let us point out that we see a main contribution of our work in a different context, as 
we have layed the fundamental building block for a conceptually correct, working variational 
framework that can combine perspective shape from shading with other techniques from
computer vision such as e.g.\ stereo vision. We aim to explore the arising possibilities in a
future work.

\bigskip

\runinhead{Acknowledgements.}
This work has been partially funded by the Deutsche For\-schungs\-gemeinschaft (DFG)
as a joint project (BR 2245/3-1, BR 4372/1-1).

\section{Appendix}
\label{sec:appendix}

\noindent 
{\bf Alternative Derivation of the Surface Normal.}
Instead of computing the derivatives with respect to the 2-D image coordinates $x$ 
and $y$, one can also derive the surface normal in an alternative way that is often used 
in the literature, see e.g. \cite{Wu_IJCV2010}. The idea is to interpret the original surface in 
Eq. \eqref{eq:77} as a function of the 3-D coordinates $X$, $Y$ and $Z(X, Y)$
\begin{equation}
\mathcal{S} \left( X( \mathbf{x}, z ), Y( \mathbf{x}, z ), Z(X( \mathbf{x}, z ), Y( \mathbf{x}, z )) \right) = 
  \left[
    \begin{array}{c}
      X( \mathbf{x}, z )\\ 
      Y( \mathbf{x}, z )\\ 
      Z(X( \mathbf{x}, z ), Y( \mathbf{x}, z )) 
    \end{array}
  \right] 
:=
  \left[
    \begin{array}{c}
      \dfrac{z \, x}{\mathtt{f}} \vspace{1ex}\\ 
      \dfrac{z \, y}{\mathtt{f}} \vspace{1ex}\\ 
      - z
    \end{array}
  \right] \, .
\end{equation}
Dropping the dependency of $X$, $Y$ and $Z(X,Y)$ on $\mathbf{x}$, $z$ and computing the partial derivatives with respect to $X$ and $Y$
via the chain rule
\[
\frac{\partial X}{\partial Y} = \frac{\partial X}{\partial x} \frac{\partial x}{\partial Y} \, ,
\qquad \qquad \qquad
\frac{\partial Y}{\partial X} = \frac{\partial Y}{\partial y} \frac{\partial y}{\partial X}
\]
then gives the tangent vectors to the surface
\begin{equation}
\label{eq:s_normal}
\mathcal{S}_{X}( \mathbf{x}, z ) = 
\left[
  \begin{array}{c}
  1\\[1mm]
  \dfrac{z_{x} y}{z+z_{x} \, x}\\[3mm]
  - \dfrac{z_{x} \, \mathtt{f}}{z+z_{x} \, x}
  \end{array}
\right]
\; ,
\;\;\;\;\;
\mathcal{S}_{Y}( \mathbf{x}, z ) = 
\left[
  \begin{array}{c}
  \dfrac{z_{y} \, x}{z+z_{y} \, y}\\[4mm]
  1\\[1mm]
  - \dfrac{z_{y} \, \mathtt{f}}{z+z_{y} \, y}
  \end{array}
\right] \; .
\end{equation}
After some computations we finally obtain the corresponding normal direction
\begin{equation}
\label{eq:normal:alt}
\hat{\mathbf{n}}( \mathbf{x} )  = 
\mathcal{S}_{X}( \mathbf{x}, z )
\times
\mathcal{S}_{Y}( \mathbf{x}, z ) 
 = 
\dfrac{\mathtt{f}^2}{(z+z_{x} \, x)(z+z_{y} \, y)}
\,\, \mathbf{n}( \mathbf{x} )
\, .
\end{equation}
where $\mathbf{n}( \mathbf{x} )$ is the normal direction from Eq. \eqref{eq:17}
As expected, both vectors only differ by scale, i.e. they have the same direction.
Hence, the corresponding normalised vectors $\mathbf{n}/|\mathbf{n}|$ and $\hat{\mathbf{n}}/|\hat{\mathbf{n}}|$ are identical. 
While this alternative derivation was not used in our paper, it helps to clarify a 
common mistake in the literature that will be explained in the following.

\medskip

\noindent
{\bf Remark.} Please note that, unlike in the orthographic case, the cross derivatives
${\partial X}/{\partial Y}$ and ${\partial Y}/{\partial X}$
do not vanish for the perspective model.
Hence, using the orthographic derivation of the normal direction 
from Horn and Brooks \cite{HB89} 

\begin{equation}
\label{eq:normalHB}
\mathbf{n_\mathrm{ortho}}(\mathbf{x}) =  
\frac{\partial}{\partial X}\!
\left[
\begin{array}{c}
X \\
Y \\
Z 
\end{array}
\right]
\times
\frac{\partial}{\partial Y}\!
\left[
\begin{array}{c}
X \\
Y \\
Z
\end{array}
\right]
=
\left[
\begin{array}{c}
1 \\
0 \\
Z_{X}
\end{array}
\right]
\times
\left[
\begin{array}{c}
0 \\
1 \\
Z_{Y}
\end{array}
\right]
=
\left[
\begin{array}{c}
-Z_{X} \\
-Z_{Y} \\
1 
\end{array}
\right]
\end{equation}
with zero cross derivatives and simply replacing the remaining partial derivatives $Z_{X}$ and $Z_{Y}$
by the corresponding expressions from \eqref{eq:s_normal}
is {\em not completely correct }
for the perspective case.
Such an approach has for instance been proposed in \cite{Wu_IJCV2010,ZYT07}.
  It actually mixes 
the orthographic and the perspective model and thus typically gives worse results in the case of strong perspective distortions.
Moreover, apart from not being completely correct, this strategy also yields significantly more complex models
that typically require auxiliary variables to be solved, see again e.g.\ \cite{Wu_IJCV2010,ZYT07}.

\end{document}